\documentclass[onefignum,onetabnum]{siamonline250211}

\usepackage{lipsum}
\usepackage{amsfonts}
\usepackage{graphicx}
\usepackage{epstopdf}
\usepackage{algorithmic}
\ifpdf
  \DeclareGraphicsExtensions{.eps,.pdf,.png,.jpg}
\else
  \DeclareGraphicsExtensions{.eps}
\fi

\usepackage{enumitem}
\setlist[enumerate]{leftmargin=.5in}
\setlist[itemize]{leftmargin=.5in}

\newsiamremark{remark}{Remark}
\newsiamremark{example}{Example}
\newsiamremark{hypothesis}{Hypothesis}
\crefname{hypothesis}{Hypothesis}{Hypotheses}
\newsiamthm{claim}{Claim}
\newsiamremark{fact}{Fact}
\crefname{fact}{Fact}{Facts}

\headers{Mechanisms of speciation in generative Diffusion Models}{P. Causin, A. Marta}

\title{Mechanisms of speciation in generative Diffusion Models: \\[1mm] a study via Lyapunov-Schmidt reduction 
}

\author{Paola Causin \and Alessio Marta\thanks{Department of  Mathematics, University of Milan, Milan, Italy 
  (\email{paola.causin, alessio.marta@unimi.it}).}
}

\usepackage{amsopn}

\newtheorem{assumption}{Assumption}

\ifpdf
\hypersetup{
  pdftitle={Mechanisms of speciation in generative Diffusion Models},
  pdfauthor={P. Causin, A. Marta}
}
\fi

\usepackage{fontawesome5}

\begin{document}

\maketitle

% REQUIRED
\begin{abstract}
During the reverse dynamics of generative diffusion models, trajectories evolve towards increasingly specificity of the data via speciation events. Recent works inspired by statistical mechanics have interpreted these events as symmetry breaking transitions related to the dynamics of the reverse process or according to information theory-related quantities. In this work we frame this mechanism by 
developing a local theory of speciation based on dynamical systems and singularity theory. Namely, we characterize a speciation event as a degeneracy of the Hessian of the noised log-density. We use a Lyapunov--Schmidt reduction on the score to single out the uniformly stable directions and yield a scalar bifurcation equation whose coefficients are explicit posterior cumulants. We show that a generic speciation is a nondegenerate fold, i.e.\ the birth of a mode--saddle pair, whereas a pitchfork requires symmetry of the data and is unfolded by any symmetry-breaking perturbation. Importantly, we show that the nature of the fold can be predicted via an explicit computation. When considering robustness with respect to perturbations in the score we find that, to first order, a score error affects a simple speciation only through its projection onto the critical direction: the speciation level shifts linearly at a fold and according to a cusp law near a symmetric configuration, while transverse errors are absorbed by the spectral gap. Experiments on Fashion-MNIST with exact and learned scores confirm these predictions and show that the speciation of a learned score is controlled by this single component of its error mediated by a measurable gain factor. 
\end{abstract}

% REQUIRED

% REQUIRED

\section{Introduction}
\label{sec:intro}

Diffusion models (DMs) are a class of generative models that synthesize data by progressively denoising a sample drawn from pure noise, reversing a stochastic process that has gradually corrupted the data~\cite{sohl2015deep,Song2019,Ho2020,Song2021}. In the accepted theoretical framework, both the forward noising process and the reverse generative dynamics are described by stochastic differential equations whose marginal densities evolve according to Fokker--Planck equations. The reverse dynamics is obtained from the forward one through Anderson's time-reversal formula~\cite{anderson1982reverse}, or, equivalently, the (same) marginals are transported by a deterministic probability-flow ODE~\cite{Song2021}. A DM is thus a non-autonomous dynamical system living in a high-dimensional state space, driven by the score, i.e. the gradient of the noised density log-marginal. 

\medskip
 
\noindent The internal dynamics of DMs is remarkably rich, and characterized by the unfolding through a sequence of qualitative transitions. Namely, during sampling, trajectories evolve to increasingly specificity of the data via \textit{speciation events}. The series of works~\cite{raya2023spontaneous,biroli2024dynamical,ambrogioni2025statistical}, inspired by statistical mechanics, has interpreted speciations as spontaneous symmetry breaking, analogous to phase transitions in equilibrium systems, and has related them to pitchfork bifurcations of the score field. In particular,~\cite{biroli2024dynamical} identifies three regimes of the reverse dynamics: an essentially Brownian regime, a regime in which the trajectory commits to one of the main classes of the data, and a final regime in which it may collapse onto a training example, the signature of memorization. Complementary viewpoints relate speciation to the geometry and spectral structure of the data manifold~\cite{ventura2025manifolds}, to the hierarchical organization of the data~\cite{sclocchi2025phase}, and to non-asymptotic critical windows in which specific features of the output are decided~\cite{li2024critical}. 

\medskip
 
\noindent Despite this progress, the theoretical picture remains incomplete in several respects. The emergence of new attractors has been interpreted almost exclusively through the pitchfork bifurcation. A pitchfork, however, is not generic: it requires an equilibrium that persists across the transition and a $\mathbb Z_2$ symmetry along the critical direction, which are usually enforced by the choice of a symmetric data distribution and of a prescribed order parameter. Real data are not exactly symmetric, and it is natural to ask which mechanism is generic. In addition, existing analyses mainly characterize when a speciation occurs, but not which components of the high-dimensional score decide its outcome. This question is central, since the score is only known through a learned approximation, and knowing which errors can alter the generated sample (and how to influence them via guidance) is very relevant.
In this work we pursue our research in the field initiated in \cite{martacausin2026}
and prosecuted in the companion paper~\cite{softmodes2026}, and we address these questions with the tools of dynamical systems and singularity theory. Namely, we characterize a speciation as a point of degeneracy of the  Hessian of the noised log-density. A Lyapunov--Schmidt reduction allows to eliminate the uniformly stable directions and reduce the equilibrium problem to a scalar bifurcation equation for the critical amplitude~\cite{GolubitskySchaeffer1985}. Since all the jets of the noised log-density can be related to posterior cumulants, the coefficients of the reduced equation are computable in closed form and we use them to verify our hypotheses on realistic datasets.

\medskip
 
\noindent Our main contributions can be summarized as follows:
\begin{itemize}
\item \emph{Density speciation versus drift confinement.} We distinguish the bifurcations of the critical set of the noised density from the loss of stability of the stagnation points of the sampler drift. The former are invariant under the rescaling relating variance-preserving and variance-exploding processes, whereas the latter depend on the sampler and on the schedule
\item \emph{Posterior characterization and reduced equation.} Via the Lyapunov--Schmidt reduction we obtain the scalar bifurcation equation along the critical direction, whose coefficients are related to posterior cumulants
\item \emph{Study of the nature of a generic speciation.} We show that the generic speciation is a nondegenerate fold, that is, the birth of a mode--saddle pair, rather than the splitting of an existing mode. The pitchfork requires an exact reflection symmetry of the data and is unfolded by any symmetry-breaking perturbation 
\item \emph{Dimensions of the critical subspace.} 
We show that under a transversality assumption,  speciations have a one-dimensional critical subspace. Higher-dimensional critical subspaces require additional structure, such as a symmetry of the data
\item \emph{Stability with respect to score perturbations.} We show that, at first order, a score error affects a simple speciation only through its scalar projection onto the critical direction; the transverse components are absorbed by the spectral gap. After the last speciation along a branch, the response becomes isotropic and bounded
\item \emph{Numerical tests on Fashion-MNIST.} Using the exact score of the empirical distribution of pairs of Fashion-MNIST classes, we confirm that the speciation of class pairs is a fold, with fold coefficients significantly different from zero, whereas the pitchfork is recovered only by symmetrizing the data; we verify the predicted barrier exponents, the cusp law and the critical read-out. With learned scores, we
observe that the speciation level is controlled by the critical read-out of the network error mediated by a gain; small networks fail to separate the classes at all, although their global error is small.
\end{itemize}

\medskip

\noindent Structure of the paper. In \Cref{sec:diffusion} we recall the forward
and reverse dynamics of DMs and introduce the notions of density speciation and drift
confinement; \Cref{sec:mech-ex} presents three toy examples from mechanics which serve as an introduction to the concepts used below; \Cref{sec:diffusion-dictionary} draws a parallel between the mechanical examples and the context of diffusion models and provides a definition of soft critical points;
\Cref{sec:posterior} introduces the relevant quantities used in the Lyapunov--Schmidt reduction and connects them to moments of the score, performing a classification of the bifurcation types;  
\Cref{sec:critical-dimension} addresses the delicate point of the dimension of the kernel of the Hessian at singularities; \Cref{sec:stability} studies the robustness of equilibria and the sensitivity of the
committed configuration to score errors at different levels of the generative process; \Cref{sec:numerical} reports the numerical experiments on Fashion-MNIST both with exact and learned score; \Cref{sec:discussion} summarizes the resulting picture and outlines open
directions. Technical proofs and details about neural networks used in the numerical experiments are collected in \Cref{sec:proofs} and \Cref{sec:NN}, respectively.

\smallskip 
\noindent The codes that reproduce the mechanical examples and the numerical experiments are available in the repository accompanying this paper. 

\section{Diffusion models and the concept of speciation}
\label{sec:diffusion}

\subsection{Noising and denoising processes}\label{sec:noising}

Let $p_0$ be a probability measure on $\mathbb{R}^d$, the data
distribution. Data are noised by according to the Ornstein--Uhlenbeck (OU) process 
\begin{equation}\label{eq:OU}
  dX_t=-\theta X_t\,dt+\sqrt{2D}\,dW_t,\qquad X_0\sim p_0,
\end{equation}
with constants $\theta,D>0$ and $W_t$ a Brownian motion. 
With $a_t:=e^{-\theta t}$ and $q_t:=\frac D\theta(1-a_t^2)$, the OU trajectories evolve as 
\begin{equation}\label{eq:OUsol}
  X_t=a_tX_0+\sqrt{q_t}\,Z,\qquad Z\sim\mathcal N(0,I_d).
\end{equation}
Conditionally on $X_0=x_0$, $X_t\sim\mathcal N(a_t x_0,q_tI_d)$; hence, for every $t>0$,
the law of $X_t$ has the density~\cite{Pavliotis14}:
\begin{equation}\label{eq:pt} p_t(x)=\int_{\mathbb{R}^d}\frac{\exp\bigl(-\|x-a_tx_0\|^2/(2q_t)\bigr)}{(2\pi q_t)^{d/2}}\,p_0(dx_0),
\end{equation}
which, whatever $p_0$, is strictly positive and $C^\infty$
on $\mathbb{R}^d\times(0,\infty)$. Moreover, $p_t$ is solution of the Fokker--Planck (FP) equation~\cite{Pavliotis14}
\begin{equation}\label{eq:FP}  \partial_tp_t=\theta\,\nabla_x\!\cdot(x\,p_t)+D\,\Delta_xp_t ,
\end{equation}
with 
$p_t$ approaching $\mathcal N(0,(D/\theta)I_d)$ as $t\to+\infty$.

\medskip

\noindent For a horizon $T>0$ and $\tau:=T-t\in[0,T]$, the process
$Y_\tau:=X_{T-\tau}$ satisfies, under mild integrability conditions on $p_0$, the Anderson's reverse SDE (R-SDE)~\cite{anderson1982reverse}
\begin{equation}\label{eq:RSDE}
  dY_\tau=\bigl[\theta Y_\tau+2D\,\nabla_x\log p_{T-\tau}(Y_\tau)\bigr]d\tau+\sqrt{2D}\,d\overline W_\tau,
  \qquad Y_0\sim p_T,
\end{equation}
where $\overline W_\tau$ is another Brownian motion 
and $\operatorname{Law}(Y_\tau)=p_{T-\tau}$.
The same marginals are transported by the deterministic
probability-flow ODE (PF-ODE)~\cite{Song2021}
\begin{equation}\label{eq:PFODE}
  \frac{dY_\tau}{d\tau}=\theta Y_\tau+D \nabla_x\log p_{T-\tau}(Y_\tau) 
\end{equation}
Both the R-SDE and PF-ODE samplers require only the score $S_t(x):= \log p_t(x)$
of the noised marginals, which in practice
is replaced by a learned approximation.

\subsubsection{Parametrization in accumulated variance} Since the theory developed in this paper concerns the noised marginals,
and not the particular sampler or noise schedule, it is convenient to reparametrize the
processes by the \emph{accumulated variance}
\begin{equation}\label{eq:clock}
  \eta(t):=\frac{q_t}{a_t^2}=\eta_\infty\bigl(e^{2\theta t}-1\bigr),\qquad \eta_\infty:=\frac D\theta ,
\end{equation}
an increasing bijection of $t\in(0,\infty)$ onto $\eta\in(0,\infty)$, with inverse 
$t(\eta) = \frac{1}{2\theta}\log\bigl(1+\eta/\eta_\infty\bigr)$. Henceforth we will use $\eta$ as the time variable and, with a slight abuse of notation,
write $a_\eta := a_{t(\eta)}$, $q_\eta := q_{t(\eta)}$, $X_\eta := X_{t(\eta)}$.
In this clock
\begin{equation}\label{eq:aq}
  a_\eta=\Bigl(\frac{\eta_\infty}{\eta_\infty+\eta}\Bigr)^{1/2},\qquad
  q_\eta=a_\eta^2\,\eta=\frac{\eta_\infty\,\eta}{\eta_\infty+\eta},
\end{equation}
and the OU trajectories in~\cref{eq:OUsol} given by
\begin{equation}\label{eq:OUsol-eta} X_\eta=a_\eta\bigl(X_0+\sqrt\eta\,Z\bigr)
\end{equation}
where $\eta$ is the variance of the Gaussian noise accumulated on the data before
the contraction by $a_\eta$, and $1/\eta=a_\eta^2/q_\eta$ is the signal-to-noise
ratio (SNR). In addition, from now on we write $p_\eta$
for the marginal at accumulated variance $\eta$, and $\varphi_\eta=
\log p_\eta$ for its logarithm, and we define the score and Hessian of $\varphi_\eta$ as 
\begin{equation}\label{eq:score}
S_\eta(x):=\nabla_x \varphi_\eta(x),\qquad H_\eta(x):=\nabla_x S_\eta(x)=\nabla_x^2 \varphi_\eta(x)
\end{equation}
Under this parametrization, the drift
term  in the samplers \cref{eq:RSDE} and \cref{eq:PFODE}  becomes
\begin{equation}\label{eq:drift}
  b_\eta(x) = \theta x + c_s D\, S_\eta(x),
  \qquad
  c_s =
  \begin{cases}
    2 & \text{for the R-SDE},\\[2pt]
    1 & \text{for the PF-ODE}
  \end{cases}
\end{equation}
which, since $d\eta/dt = 2D/a_\eta^2$, yields
the forms
  \begin{align}      
\text{R-SDE}: \quad  & dY_\eta = -\frac{a_\eta^2}{2D}\, b_\eta(Y_\eta)\, d\eta + a_\eta\, d\overline W_\eta
  \\[2mm]
\text{PF-ODE}: \quad   & \frac{dY_\eta}{d\eta} = -\frac{a_\eta^2}{2D}\, b_\eta(Y_\eta)
 \end{align}
where $\overline W$ is a Brownian motion running backwards in $\eta$, and the noise amplitude
follows from $\sqrt{2D}\,\sqrt{|d\tau|} = a_\eta \sqrt{|d\eta|}$.

\medskip

\noindent
The physical time $t$ will be retained
only in the numerical experiments of \Cref{sec:numerical}.

\subsection{Density speciation and drift confinement}
\noindent We denote the critical set of the density at noise level $\eta$ as 
\begin{equation}
\Xi_\eta=
\{x \in \mathbb R^d \, : \, \nabla_x p_\eta(x) = 0 \} \equiv 
\{x \in \mathbb R^d \, : \, S_\eta(x)=0\}
\end{equation}
We call \textit{density speciation} a local bifurcation of the critical points of \(p_\eta\) that changes the number of its local maxima. At a bifurcation point $(x_c,\eta_c)$,
\begin{equation}
S_c:=S_{\eta_c}(x_c)=0, \quad 
\text{and} \quad H_c:=H_{\eta_c}(x_c) \quad 
\text{is singular}
\end{equation}
% In the corank-one setting considered below, \(\ker H_c=\operatorname{span}\{e_c\}\), while \(H_c\) is negative definite on \(e_c^\perp\). This spectral condition identifies a single critical mode and stable transverse directions; the actual change in the critical-point structure is established by the nondegeneracy conditions on the reduced equation (see \Cref{sec:LS}).
The zeros of the drift $b_\eta$, which we call \emph{stagnation points}, are the fixed points
\begin{equation}\label{eq:stagnation}
  x=-c_s\,\eta_\infty\,S_\eta(x),
\end{equation}
and are therefore distinct from the critical points of the density.
% We can however establish 
% a correspondence of speciation events in 
% processes with and without drift. 
% Let $\widehat X_\eta=X_0+\sqrt\eta\,Z$ be a pure Brownian motion run on the clock $\eta$ and $\widehat S_\eta:=\nabla_x\log\widehat p_\eta$ and $\widehat H_\eta:=\nabla_x^2\log\widehat p_\eta$ be the related score and Hessian.
% \begin{proposition}[Equivalence of the score landscapes]\label{prop:OUVE}
% Let $x\in \mathbb R^d$. 
% For every $\eta>0$: (i) $x^\ast$ is a zero of $S_\eta$ if and only if $\widehat x^\ast=x^\ast/a_\eta$ is a zero of $\widehat S_\eta$, and $x^\ast\mapsto \widehat x^\ast$ is a bijection of the critical sets that preserves their connected components and degeneracies; (ii) $H_\eta(x^\ast)$ and $\widehat H_\eta(\widehat x^\ast)$ have the same eigenvectors, with eigenvalues related by $\lambda=a_\eta^{-2}\widehat\lambda$.
% \end{proposition}
% \begin{proof}
%  By \eqref{eq:OUsol-eta}, $X_\eta=a_\eta\widehat X_\eta$  and we have the relations
% \begin{equation}\label{eq:OU-VE}
%   p_\eta(x)=a_\eta^{-d}\,\widehat p_\eta(x/a_\eta),\qquad
%   S_\eta(x)=a_\eta^{-1}\widehat S_\eta(x/a_\eta),\qquad
%   H_\eta(x)=a_\eta^{-2}\widehat H_\eta(x/a_\eta)
% \end{equation}
%    \end{proof}
%%
We can however establish a relation with  driftless processes by observing that 
the OU dynamics admits the explicit solution
$X_t = e^{-\theta t}X_0 + \sqrt{2D}\int_0^t e^{-\theta(t-s)}\,dW_s$, which, dividing by
$a_t = e^{-\theta t}$, gives $X_t/a_t = X_0 + M_t$, where
$M_t := \sqrt{2D}\int_0^t e^{\theta s}\,dW_s$ is a continuous Gaussian martingale with
\begin{equation*}
  \langle M\rangle_t = 2D\int_0^t e^{2\theta s}\,ds = \eta(t).
\end{equation*}
Hence $B_\eta := M_{t(\eta)}$ is a standard Brownian motion, and
\begin{equation}\label{eq:VE-pathwise}
  X_\eta = a_\eta \widehat X_\eta, \qquad \widehat X_\eta := X_0 + B_\eta ,
\end{equation}
Thus, up to the deterministic rescaling $a_\eta$, the forward process is a variance-exploding
process driven by the same noise. For fixed $\eta$,
$\widehat X_\eta = X_0 + \sqrt{\eta}\,Z$, consistently with \eqref{eq:OUsol-eta}.

\begin{proposition}[Equivalence of the score landscapes]\label{prop:equivalence}
For every $\eta > 0$:
\begin{enumerate}[label=(\roman*)]
  \item $x^*\in\mathbb R^d$ is a zero of $S_\eta$ if and only if
        $\hat x^* := x^*/a_\eta$ is a zero of $\widehat S_\eta$, and
        $x^*\mapsto \hat x^*$ is a bijection of the critical sets that preserves their
        connected components and degeneracies;
  \item $H_\eta(x^*)$ and $\widehat H_\eta(\hat x^*)$ have the same eigenvectors, with
        eigenvalues related by $\lambda = a_\eta^{-2}\widehat\lambda$.
\end{enumerate}
\end{proposition}

\begin{proof}
Only the law of $X_\eta$ at fixed $\eta$ matters here. Since
$X_\eta \overset{d}{=} a_\eta \widehat X_\eta$, the linear change of variables
$x = a_\eta z$ gives
\begin{equation}\label{eq:2.17}
  p_\eta(x) = a_\eta^{-d}\,\widehat p_\eta(x/a_\eta), \qquad
  S_\eta(x) = a_\eta^{-1}\,\widehat S_\eta(x/a_\eta), \qquad
  H_\eta(x) = a_\eta^{-2}\,\widehat H_\eta(x/a_\eta).
\end{equation}
Since $a_\eta > 0$, the second relation yields (i): the map $x\mapsto x/a_\eta$ is a
linear homeomorphism of $\mathbb R^d$, and therefore it preserves connected
components. The third relation yields (ii), and in particular
$\operatorname{rank} H_\eta(x^*) = \operatorname{rank} \widehat H_\eta(\hat x^*)$,
so degeneracies are preserved as well.
\end{proof}

\begin{example}[Symmetric doublet: speciation versus confinement]\label{ex:doubletOU}
Let $d=1$ and $p_0(x)=\tfrac12(\delta_m(x)+\delta_{-m}(x))$, $m>0$. Setting
$\mu_\eta:=a_\eta m$, the marginal is
$p_\eta=\tfrac12\bigl(\mathcal N(\mu_\eta,q_\eta)+\mathcal N(-\mu_\eta,q_\eta)\bigr)$, with score
\begin{equation}\label{eq:doublet-score}
  S_\eta(x)=\frac{1}{q_\eta}\Bigl(-x+\mu_\eta\tanh\frac{\mu_\eta x}{q_\eta}\Bigr).
\end{equation}
By symmetry the origin is a critical point for every $\eta$, and
\begin{equation}\label{eq:doublet-H}
  H_\eta(0)=\frac{1}{q_\eta}\Bigl(\frac{\mu_\eta^2}{q_\eta}-1\Bigr)
  =\frac{(m^2-\eta)(\eta+\eta_\infty)}{\eta_\infty\,\eta^2}.
\end{equation}
A density speciation event occurs
at $\eta_c=m^2$, which separates the regime
in which the origin is a maximum of $p_\eta$ ($\eta>m^2$) from the one where the origin is a minimum ($\eta<m^2$).
In the drift-less frame, $\widehat p_\eta
= \tfrac12\bigl(\mathcal N(m,\eta) + \mathcal N(-m,\eta)\bigr)$ and the score and Hessian are
\begin{equation}\label{eq:doublet-VE}
  \widehat S_\eta(z) = \frac{1}{\eta}\Bigl(-z + m\tanh\frac{mz}{\eta}\Bigr),
  \qquad
  \widehat H_\eta(z) = \frac{1}{\eta}\Bigl(\frac{m^2}{\eta}\,
  \operatorname{sech}^2\frac{mz}{\eta} - 1\Bigr),
\end{equation}
so that
\begin{equation}\label{eq:doublet-VE-origin}
  \widehat H_\eta(0) = \frac{m^2-\eta}{\eta^2}
\end{equation}
and the speciation threshold is again
$\eta^* = m^2$, the same in both frames. 
The origin is also a stagnation point of the drift, since
$S_\eta(0) = 0$ gives $b_\eta(0) = 0$.
Linearizing in space the sampler $dY/d\tau = b_\eta(Y)$
around it, a small deviation $Y = \delta$ for small $\delta$, obeys $d\delta/d\tau = b'_\eta(0)\,\delta$, so
\begin{equation}\label{eq:drift-linearization}
  \delta(\tau) = \delta(\tau_0)\exp\Bigl(\int_{\tau_0}^{\tau} b'_{\eta(s)}(0)\,ds\Bigr),
  \qquad
  b'_\eta(0) = \theta + c_s D\,H_\eta(0).
\end{equation}
Since the dynamics is non-autonomous, the sign of $b'_\eta(0)$ does not determine
stability in the classical sense. It tells whether nearby trajectories are
instantaneously contracted ($b'_\eta(0)<0$) or expanded ($b'_\eta(0)>0$) towards or
away from the origin. Because $\eta_\infty = D/\theta$, the transition occurs at
$H_\eta(0) = -1/(c_s\eta_\infty)$ and not at $H_\eta(0) = 0$. Namely, the transition happens when 
\begin{equation}\label{eq:2.23}
  c_s\,(\eta - m^2)(\eta + \eta_\infty) = \eta^2 
\end{equation}
Its solution $\eta_s$ satisfies $\eta_s > m^2 = \eta^*$, and it depends on both the
sampler and the noise schedule:
$\eta_s = m^2\eta_\infty/(\eta_\infty - m^2)$ for the PF-ODE
(provided $\eta_\infty > m^2$), and
$\eta_s = \sqrt{\eta_\infty^2 + m^4} - (\eta_\infty - m^2)$ for the R-SDE.
The origin thus becomes repelling for the drift before speciation occurs in the
density. This threshold has no counterpart in the driftless frame.
The two confinement thresholds are ordered as
\begin{equation}\label{eq:doublet-order} \eta_c<\eta_{\mathrm{conf}}^{\mathrm{R\text{-}SDE}}<\eta_{\mathrm{conf}}^{\mathrm{PF}}
\end{equation}
In both cases, for
$\eta\in(\eta_c,\eta_{\mathrm{conf}})$ the density is still unimodal,
$H_\eta(0)<0$, but trajectories start to diverge from
the origin.
In \cref{fig:symdoub} we show the trajectories 
for the PF-ODE sampler in a OU process
and in the driftless counterpart.
\begin{figure}
    \centering    \includegraphics[width=0.95\linewidth]{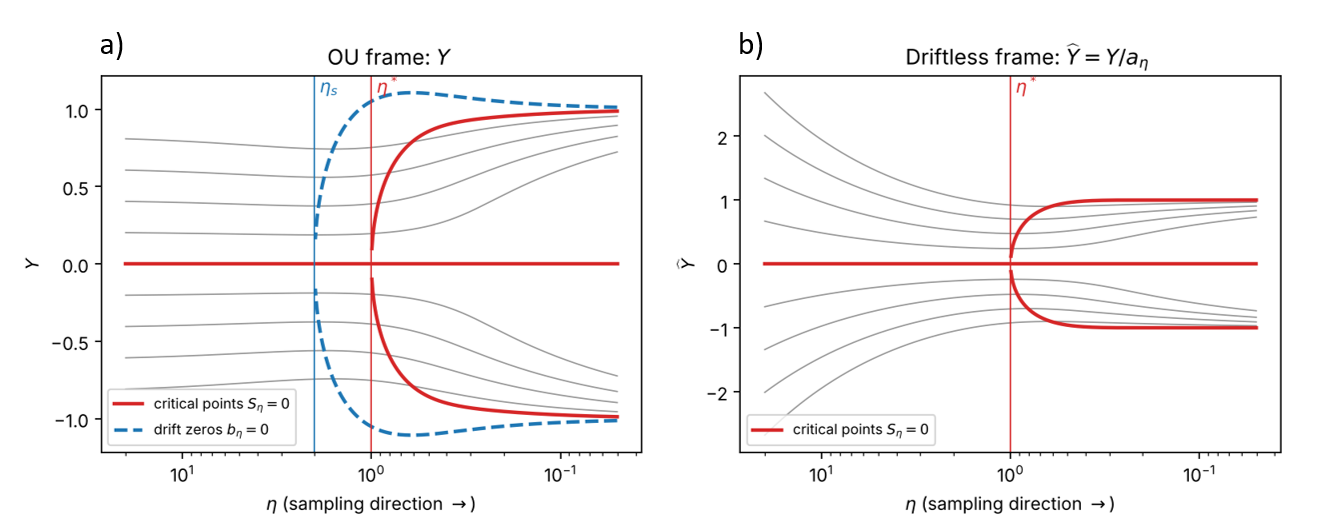}
    \caption{Symmetric doublet with $m=1$, $\theta=1$, $D=2$, sampler: PF-ODE.
(a) OU frame: drift zeros (dashed) bifurcate at $\eta_s$, critical points of the score
(solid) at $\eta^*$; grey curves are sampler trajectories.
(b) Driftless frame $\widehat Y = Y/a_\eta$: bifurcation of the density occurs again at $\eta^*$.}
    \label{fig:symdoub}
\end{figure}

\end{example}

%%%%%% PECLET %%%%%%%%%%%%%%
% \subsection{Nondimensional forms and P\'eclet number}
% \paola{spostare}
%  Introducing the characteristic length $L$ and diffusive time $D/L^2$, we get the nondimensional variables $\hat x=x/L$ and 
% $\hat t = tD/L^2$, so that the FP can be rewritten as
% \begin{equation}
% \partial_{\hat t} \, p = \Pe_0 \partial_{\hat x}(\hat x\,p)
% + \partial_{\hat x\hat x} p,
% \label{eq:ou-nd}
% \end{equation}
% where the non-dimensionalization procedure has made appear the ``structural'' P\'eclet number
% $\Pe_0 = \frac{\theta L^2}{D}$
% which measures the relative importance of drift and diffusion. 
% In dimensionless variables the total drift reads 
% \begin{equation}
% \hat u = -\Pe_0 \hat x - \hat S,
% \label{eq:total-drift-nd}
% \end{equation}
% with $\hat S=\partial_{\hat x}\log p$,
%%%%%% PECLET %%%%%%%%%%%%%%

% For each fixed diffusion time $t$, we measure the score approximation error through the Fisher divergence
% weighted by the data distribution
% \begin{equation}
%   \mathcal{D}_t(S_\theta\Vert S_t;p_t) \;=\; \E_{x\sim p}\bigl\|S_\theta(x)-S_t(x)\bigr\|^2 = 
%  \int ||S_\theta(x)-S_t(x)||^2p_t(x)dx,
%   \label{eq:fisher}
% \end{equation}
% equivalently $\|S_\theta-S_t\|_{L^2(p_t)}^2$. 

% \medskip
% \noindent In the following we focus thus on the study of the density mode speciation. 

\section{Lyapunov-Schmidt reduction: introductory examples}
\label{sec:mech-ex}
Before discussing the reduction framework for diffusion models which is the core of the paper, we present three
mechanical examples, elaborated from \cite[Ch.~1]{GolubitskySchaeffer1985}, which 
isolate the main mechanisms used throughout the paper and directly translate to the problem of our interest. 
Although this material will be familiar to experts in bifurcation theory~\cite{zbMATH01552061}, we include it for the benefit of a broader readership, in particular from the machine-learning community, for whom it may be less standard.
The first system exhibits
a symmetry-induced pitchfork bifurcation: a stable equilibrium loses stability
and gives rise to two symmetry-related branches. The second example shows how
small symmetry-breaking perturbations on the same system unfold the ideal pitchfork bifurcation. Finally,
the three-rod system introduces a multidimensional state space and illustrates
the role of the Lyapunov--Schmidt (LS) reduction: the transverse variables are slaved
to a critical soft mode, and the equilibrium problem reduces locally to a
scalar normal form.
The codes that compute the bifurcation diagrams and integrate the dynamics are provided in the \texttt{MechExamples} folder of the repository accompanying this paper.

\medskip

\begin{figure}[ht]
\centering
\includegraphics[width=\textwidth]{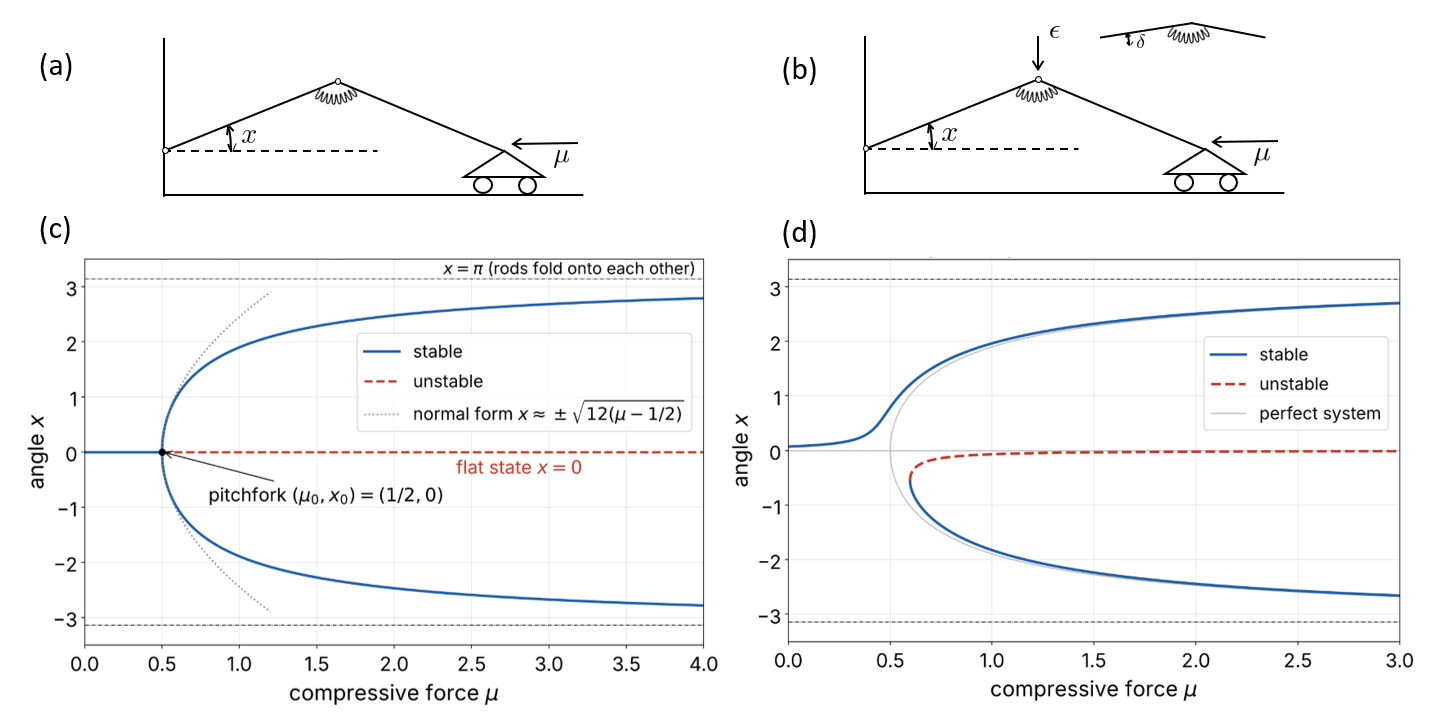} 
\caption{(a): simple mechanical system (discrete analogue of Euler buckling), which exhibits a  pitchfork bifurcation (c) at $(x_0,\mu_0)=(0,\frac12)$;  
(b): mechanical system with perturbations $\delta=0.1, \epsilon=0.03$ in the describing model and (d) corresponding altered bifurcation diagram with unfolding of the pitchfork .}
\label{fig:mechex12}
\end{figure}

\paragraph{Two-rod system: symmetric pitchfork} Consider the simple mechanical system shown in~\cref{fig:mechex12}(a),
which consists of two rigid rods of unit length, 
connected by pin-like hinges, and 
subjected to
a compressive force $\mu$ resisted by a torsional spring 
of unit strength. The state of the system is described by one degree of freedom, the angle $x$ (with $|x|<\pi$) measuring the deviation from the horizontal resting position. 
The potential energy of the system  is
\begin{equation}
V_\mu(x)=\frac{x^2}{2}+2\mu
(\cos x-1)
\end{equation}
and steady--states are the  solutions of
\begin{equation}
0=F_\mu(x)=-D_x V_\mu(x)=-x+2\mu \sin x
\label{eq:stcarr}
\end{equation}
A singularity appears at  $(x_0,\mu_0)=\left(0,\frac12\right)$, where also 
$D_x F_{\mu}(x_0)=0$.
As $\mu$ crosses the value $\mu_0$,
the number of zeros of $F_\mu$
jumps from one (resting state) to three (resting state now unstable and rods buckled up/down),
see also the codes and the bifurcation diagram  in~\cref{fig:mechex12}(c).

\smallskip
\noindent 
We expand $F_\mu$ in $(x,\mu)$ near the bifurcation point as 
\begin{equation}
\begin{aligned}\label{eq:tworod-expansion}
  F_\mu(x) \approx & F_{\mu_0}(x_0)+
  D_x F_{\mu_0}(x_0) (x-x_0) + 
  D_\mu F_{\mu_0}(x_0) (\mu-\mu_0) +  D_{xx} F_{\mu_0}(x_0) \frac{(x-x_0)^2}{2} + \\[1mm]
&
  D_{x}D_{\mu} F_{\mu_0}(x_0)(x-x_0)(\mu-
  \mu_0) + D_{\mu\mu} F_{\mu_0}(x_0) \frac{(\mu-\mu_0)^2}{2}
  + 
  D_{xxx} F_{\mu_0}(x_0) \frac{(x-x_0)^3}{6} \\[1mm]
 &  = 2x\left(\mu-\frac12\right) - \frac16 x^3,
\end{aligned}
\end{equation}
where to obtain the form in the last row we have observed that the only nonzero terms are $D_xD_\mu F_\mu=2\cos x_0=2$ and $D_{xxx}F_\mu=-2\mu_0\cos x_0=-1$. Setting $\kappa=2(\mu-\frac12)$, $c_3=\frac16$, $u=x$ and $\Phi_\kappa \equiv F_\mu$, \eqref{eq:tworod-expansion} gives, to leading order, the supercritical pitchfork
\begin{equation}\label{eq:pitchfork-nf}
  \Phi_\kappa(u)=\kappa u-c_3u^3,
\end{equation}
whose nontrivial branches are $u=\pm\sqrt{\kappa/c_3}=\pm\sqrt{6\kappa}$, that is, $x\approx\pm\sqrt{12(\mu-\frac12)}$. 
The study of the stability of the solutions would require to consider the time evolution of the system, but here we limit ourselves to notice that $D_xF_\mu(x)=
2\mu \cos x-1$, so that for $\mu<\frac 12,
D_x F_\mu(0)<0$ and $x=0$ is stable,
while for $\mu>\frac 12$ it is unstable. 
The two branches emerging 
for $\mu> \frac12$ are both stable since, with $\mu^*=x^*/(2\sin x^*)$, 
$D_x F_{\mu^*}(x^*) = x^*\cot x^* -1< 0$ since $|x|<\pi$.

\medskip

\paragraph{Two-rod system with perturbations in the model: fold} Having established the above 
framework, 
we pass to study how
perturbations of the model (in the sense specified below of symmetry-breaking perturbations) impact the bifurcation diagram. To do this
we add to the system a small vertical force $\epsilon$
and we consider a slightly asymmetric torsional spring,
which has rest angle 
$x=\delta$, and not $x=0$
(see \cref{fig:mechex12}(b)).
The corresponding potential is 
\begin{equation}
V_\mu(x;\epsilon, \delta)=
\frac{(x-\delta)^2}{2}+
2\mu(\cos x-1)+\epsilon \sin x
\end{equation}
and the steady states are solution of 
\begin{align}
0&=F_\mu(x; \epsilon,\delta) =   -D_x V_\mu(x; \epsilon,\delta)=\delta -x +2\mu\sin x - \epsilon\cos x
\label{eq:ex-pert}
\end{align}
For fixed \((\epsilon,\delta)\), a singularity appears at point $(x_0,\mu_0)$, solution 
of~\cref{eq:ex-pert} and such that $D_x F_\mu(x;\epsilon,\delta)=0$, with coordinates:  
$$
\mu_0=\frac{1-\epsilon\sin x_0}{2\cos x_0}, \quad x_0: \quad 
(x_0-\delta)\cos x_0-\sin x_0+\epsilon=0.
$$
Proceeding as in the first example,  we expand again $F_\mu$ 
in $(x,\mu)$ about the bifurcation point
as in \cref{eq:tworod-expansion}.
Observing that, at $\mu=\mu_0$, 
\(F_{\mu}(x_0)-(\delta-\epsilon)=0, D_xF_{\mu}(x_0)=D_\mu F_{\mu}(x_0)=0\), 
and that
 $D_x D_\mu F_{\mu}(x_0)=2, D_{xx}F_{\mu}(x_0)=\epsilon$,
and $D_{xxx}F_{\mu}(x_0)=-1$,
 we obtain 
\begin{equation}
\begin{aligned}
    F_\mu(x) \approx   
(\delta - \epsilon) +2\left(\mu -\frac12\right )x 
+\frac{\epsilon}{2}x^2-
\frac{1}{6}x^3
\label{eq:F-pert-expa-1}
\end{aligned}
\end{equation}
which, setting $u=x$, $\Phi \equiv 
6F$ and $\kappa:=12(\mu-\tfrac12)$,  
has the \textit{universal unfolding form} 
\begin{equation}
\Phi_\kappa(u; \alpha)=
\alpha_1 + \kappa u  + \alpha_2 u^2
- c_3 u^3,
\label{eq:phifold}
\end{equation}
where $\alpha=(\alpha_1,\dots, 
\alpha_k) \in \mathbb R^k$ is the unfolding parameter vector (in our example
$\alpha=(\alpha_1,\alpha_2) \in \mathbb R^2$ is obtained from the linear map  $(\epsilon,\delta) \to (\alpha_1,\alpha_2)=(6(\delta-\epsilon),3\epsilon)$). Notice that
for $\alpha \equiv 0$ we recover 
the pitchfork form~\cref{eq:pitchfork-nf} of the previous example.

\smallskip

\noindent For the parameter values used in~\cref{fig:mechex12}(d), $\delta=0.1, \epsilon=0.03$, a fold occurs at approximately \((x_0,\mu_0)=(-0.554,0.597)\), where $D_\mu F_{\mu_0}(x_0)=2\sin(x_0)$,
$D_{xx}F_{\mu_0}(x_0)=-2\mu_0 \sin(x_0)+ \epsilon \cos(x_0)$
are both nonzero. The bifurcation diagram presents thus a fold with a continuous branch
starting from  $x \approx \delta -
\epsilon$ which gradually deviates toward a preferred side depending on the sign of $\delta-\epsilon$ and  
a disjoint hairpin branch which forms at $(x_0,\mu_0)$. 
In this situations, for the scalar relaxation \(\dot x=F_\mu(x; \epsilon,\delta)\), at fixed \(\mu\), the signs of \(D_xF_\mu\) at consecutive simple zeros alternate~\cite{Strogatz2015}. In \cref{fig:mechex12}(d), the continuous branch is stable and the inner and outer arms of the hairpin are unstable and stable, respectively. 
Notice how away from the degenerate points the pitchfork and perturbed diagrams look alike. 

\medskip

\begin{figure}[ht]
\centering
\includegraphics[width=.92\textwidth]{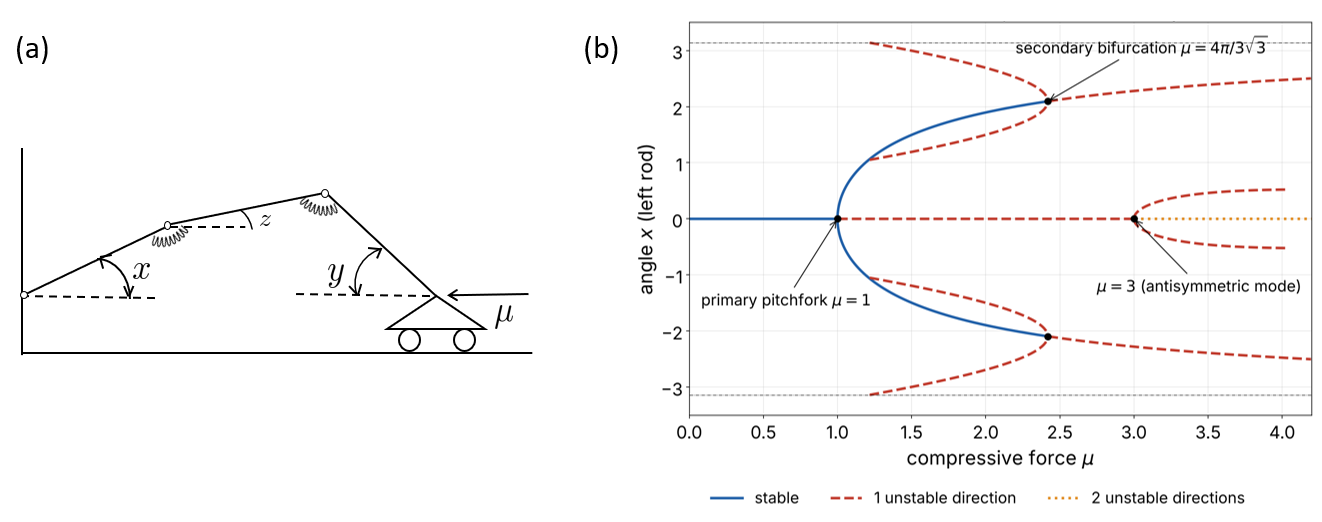} 
\caption{(a): Three--rod mechanical system (2 dofs); (b): bifurcation diagram obtained after Lyapunov-Schmidt reduction. 
The diagram reports for completeness the full numerical study of all the bifurcation points, while in the text we have limited our discussion to the simplest case of the bifurcation at $\mu=1$.}
\label{fig:mechex3}
\end{figure}

\paragraph{Three-rod system: Lyapunov-Schmidt reduction} Finally, we consider a more complex mechanical system consisting of three rods of unit length connected by pin hinges (see \Cref{fig:mechex3}(a)) and two torsional springs of unit strength.  From the geometric constraint $\sin x+\sin z=\sin y$, the system has two degrees of freedom,
$x$ and $y$. We set $w=(x,y)^\top$.
 The potential is
\begin{equation}
V_\mu(w) = \tfrac12(x-z)^2 + \tfrac12(y+z)^2 + \mu(\cos x + \cos z  + \cos y-3)
\end{equation}
and, observing that 
$z=(z_x,z_y)^\top=\left(-\frac{\cos x}{\cos z}, \frac{\cos y}{\cos z}\right)^\top$,
the steady states of the system are
\begin{equation}
0=F_\mu(w)=-D_w V_\mu(w) = 
-\left(
\begin{array}{c}
(x-z) - \mu\sin x + z_x\big((y+z) - (x-z) - \mu \sin z\big) \\[1mm]
(y+z) - \mu\sin y + z_y\big((y+z) - (x-z) - \mu\sin z\big)
\end{array}
\right)
\label{eq:3rods}
\end{equation}
The origin $w_0=(0,0)^\top$ is an equilibrium for every $\mu$. Since
$z=y-x+O(|w|^3)$ near $w_0$, the Jacobian of $F_\mu$ at the origin (that is, minus
the Hessian of $V_\mu$) is
\begin{equation}\label{eq:H0rod}
  H_0:=D_wF_\mu(w_0)=\begin{pmatrix}2\mu-5&4-\mu\\4-\mu&2\mu-5\end{pmatrix},
\end{equation}
and it admits eigenvectors $e_c=(1,1)^\top/\sqrt2$, $e_s=(1,-1)^\top/\sqrt2$ and eigenvalues
$\lambda_c=\mu-1$, $\lambda_s=3\mu-9$, respectively. At $\mu_0=1$ we have
$\lambda_c=0$, while $\lambda_s=-6<0$, and $\lambda_s$ stays negative in a
neighborhood of $\mu_0$. Hence $\operatorname{dim} \ker H_0=1$ and $\ker H_0=\operatorname{span}\{e_c\}$ at $\mu_0$. We call
$e_c$ the \emph{critical mode} and $e_s$, which spans its orthogonal complement,
the \emph{transverse mode}: on $\operatorname{span}\{e_s\}$ the matrix $H_0$ is invertible
and uniformly stable. The critical mode corresponds to $x=y$, hence to $z=0$
(symmetric buckling), whereas $e_s$ corresponds to $x=-y$ (antisymmetric
deformation). Since $H_0$ is symmetric, $\{e_c,e_s\}$ is an orthonormal basis and
we can write
\begin{equation}\label{eq:ab}
  w=a\,e_c+b\,e_s=\frac{1}{\sqrt2}\begin{pmatrix}a+b\\a-b\end{pmatrix},\qquad
  a=e_c^\top w=\frac{x+y}{\sqrt2},\qquad b=e_s^\top w=\frac{x-y}{\sqrt2},
\end{equation}
where $a$ and $b$ are the critical and transverse amplitudes. Projecting the
equilibrium equation onto the two directions gives the scalar components
\begin{align}
  \text{critical:}&\quad G^c_\mu(a,b):=e_c^\top F_\mu(a\,e_c+b\,e_s), \label{eq:Gc}\\
  \text{transverse:}&\quad G^s_\mu(a,b):=e_s^\top F_\mu(a\,e_c+b\,e_s), \label{eq:Gs}
\end{align}
and $F_\mu(w)=0$ is equivalent to $G^c_\mu(a,b)=G^s_\mu(a,b)=0$.

\medskip

\noindent \emph{Slaving of the transverse amplitude.} We have $G^s_\mu(0,0)=0$ for every $\mu$ and $\partial_bG^s_{\mu_0}(0,0)=e_s^\top H_0e_s=\lambda_s=-6\neq0$ at $\mu_0=1$. Under this latter condition, the Implicit Function Theorem (IFT), applied to the transverse equation $G^s_\mu(a,b)=0$ with $(a,\mu)$ as parameters, gives a neighborhood of $(a_0,\mu_0)=(0,1)$ and a unique smooth function $b=B_\mu(a)$ defined on it, such that
\begin{equation}
  G^s_\mu\bigl(a,B_\mu(a)\bigr)=0\qquad\text{for all $(a,\mu)$ in the neighbourhood},
\end{equation}
with $B_\mu(0)=0$ for every $\mu$, by uniqueness. 
Locally, the equilibria are in one-to-one correspondence with the zeros of the reduced scalar function
\begin{equation}\label{eq:ghat-def}
  \widehat G_\mu^c(a):=G^c_\mu\bigl(a,B_\mu(a)\bigr).
\end{equation}
Notice that, differentiating $G^s_\mu(a,B_\mu(a))\equiv0$ in $a$ gives $\partial_aG^s_\mu+\partial_bG^s_\mu\,\partial_aB_\mu=0$, which at $a_0=0$ reads 
$e_c^\top D_wF_\mu\,e_c+ e_c^\top D_wF_\mu\,e_s B_\mu(a)=e_c^\top \lambda_c
e_c+ e_c^\top \lambda_s
e_s B_\mu(a)=0$, hence $\partial_aB_\mu(0)=0$. This latter combined with $B_\mu(0)=0$ 
gives for this example the trivial slaving $B_\mu(a)\equiv 0$ (but for a general case the slaving is  non-trivial!).

\smallskip

\noindent We  now proceed to expand $\widehat G_\mu^c(a)$ in $(a,\mu)$ near the bifurcation, following the lines of the previous examples. Observing that  
\begin{equation}
    \begin{aligned}
& \widehat G_\mu^c(0)=0, \qquad  \partial_a\widehat G_\mu^c(0)=0 \\ 
&  \partial_\mu\widehat G_\mu^c(0)=0,  
 \qquad \partial_{aa}\widehat G^c_\mu(0) =0,\\
 &  \partial_a\partial_\mu\widehat G^c_\mu(0)=1, 
\qquad \partial_{aaa}\widehat G^c_\mu(0) =-\frac12  
\end{aligned}
\end{equation}
we get at leading order
\begin{equation}\label{eq:ghat}
  \widehat G_\mu(a) \approx (\mu-1)a -\frac{1}{12}a^3 
\end{equation}
which is again the pitchfork normal form~\eqref{eq:pitchfork-nf} with $u=a$, $\Phi=\widehat G_\mu(a)$ and $\kappa=(\mu-1)$,
$c_3=1/12$. Notice that linear term describes the crossing with respect
to the parameter, while the cubic term 
determines the bifurcating branches $a_\pm\approx\pm\sqrt{12(\mu-1)}$, that is,
$x=y\approx\pm\sqrt{6(\mu-1)}$ (see~\cref{fig:mechex3}(b)). 
Under the dynamics $\dot w=F_\mu(w)=-D_wV_\mu(w)$, the bifurcation at $\mu=1$ is a supercritical pitchfork
with exchange of stability. For $\mu<1$ the symmetric equilibrium is
asymptotically stable, since $\partial_a\widehat G^c_\mu(0)=\mu-1<0$ and $\lambda_s<0$. As
$\mu$ crosses $1$, the origin loses stability along the critical direction but
persists as a saddle, while two symmetry-related stable branches are created,
along which $\partial_a\widehat G_\mu^c(a_\pm)\approx -2(\mu-1)+O(\mu^2)<0$.

\begin{remark}\label{rem:threerod-fold}
More generally, when the symmetry of the full system is broken, the reduced
equation acquires imperfection terms that lead to folding events, as in the
two-rod system with perturbations form~\cref{eq:phifold}.
\end{remark}

\section{Speciation as a bifurcation of the noised density}
\label{sec:diffusion-dictionary}
We now enter in the core of the original research work and we adapt and extend the notions introduced in \Cref{sec:mech-ex} to diffusion models.
Since, by~\Cref{prop:equivalence}, density speciation is invariant under the OU--VE rescaling, we directly work in the following with the VE process parameterized by its accumulated variance \(\eta\). For ease of notation, we will drop the hat symbol on the quantities. 

\subsection{A dictionary for diffusion models}

Before proceeding, we summarize in~\cref{tab:dict} the correspondence between the quantities of the mechanical examples and their counterparts for diffusion models. The table is meant as a practical reference to connect concepts in the toy examples with the more complex problem at hand. Notice that some diffusion-side quantities are defined precisely in the following sections.
\begin{table}[h]
\centering\small
\renewcommand{\arraystretch}{1.15}
\begin{tabular}{|l|c||l|c|}
\hline
\multicolumn{2}{|c||}{\emph{Mechanical examples}} & \multicolumn{2}{c|}{\emph{Diffusion models}}\\
\hline
dofs (rod angles) & $x$, $w=(x,y)$ & noisy sample & $x\in\mathbb{R}^d$\\
load & $\mu$ & accumulated variance & $\eta$\\
buckling & $\mu \uparrow$ & reverse process & $\eta \downarrow$ \\
potential & $-V_\mu$ & log-density of marginals & $\varphi_\eta=\log p_\eta$\\
gradient field & $F_\mu=-\nabla V_\mu$ & score & $S_\eta=\nabla_x\varphi_\eta$\\
Jacobian & $D_xF_\mu$ & Hessian & $H_\eta=\nabla_x^2\varphi_\eta$\\
equilibria & $F_\mu(x)=0$ & density critical points & $S_\eta(x)=0$\\
gradient flow at fixed load & $\dot x=F_\mu(x)$ & frozen score flow & $\dot x=S_\eta(x)$\\ 
bifurcation point & $(x_0,\mu_0)$ & soft critical point & $(x_c,\eta_c)$\\
symmetry of the structure & $x\mapsto-x$, $x\leftrightarrow y$ & symmetry of the data & $R_\#p_0=p_0$\\
imperfections & $\epsilon$, $\delta$ & data asymmetry, score error & $\delta S$\\
\hline
\end{tabular}
\caption{Correspondence between quantities in the mechanical examples of~\Cref{sec:mech-ex} (left) and diffusion models (right).}
\label{tab:dict}
\end{table}

\subsection{Soft critical points and density speciation}
\begin{definition}
\label{def:soft}
We say that $(x_c,\eta_c)$, with $x_c\in\Xi_{\eta_c}$, is a \textit{soft critical
point} if $H_c:=H_{\eta_c}(x_c)$ is singular and negative semidefinite, i.e.\ all
its nonzero eigenvalues are strictly negative.
\end{definition}
A soft critical point is a density speciation if the number of local
maxima of $p_\eta$ near $x_c$ increases  as $\eta$ crosses $\eta_c$ moving towards $\eta \to 0$. Precisely,
there must exist a neighborhood $\mathcal U$ of $x_c$ and $\varepsilon>0$
such that the number of local maxima of $p_\eta$ in $\mathcal U$ is constant
on $(\eta_c-\varepsilon,\eta_c)$, constant on $(\eta_c,\eta_c+\varepsilon)$,
and different on the two intervals.

% The transverse directions are strictly stable at the point $(x_b,\eta_b)$
% only. Requiring this stability to hold uniformly on a noise window around
% $\eta_b$ is a separate hypothesis (Assumption~\ref{ass:gap}), which is not
% implied by Definition~\ref{def:speciation}.

\begin{remark}[Necessity and insufficiency of the spectral condition]
\label{rem:soft-necessary}
Negative semidefiniteness is necessary for a change in the number of maxima.
Indeed, let $(x_c,\eta_c)$ be a zero of the score and suppose $H_c$ has a positive
eigenvalue. By continuity, $H_\eta(x)$ has a positive eigenvalue for all
$(x,\eta)$ in a neighborhood of $(x_c,\eta_c)$. Hence $p_\eta$ has no local
maximum near $x_c$ for $\eta$ close to $\eta_c$, and no maxima are created
or destroyed there. This condition is however not sufficient. Indeed, a soft critical point
may persist as a degenerate maximum, or the critical eigenvalue may touch
zero without crossing it, with no change in the critical-point structure.
Whether the number of maxima actually changes is decided by the
nondegeneracy and transversality conditions on the reduced bifurcation
equation. This is the object of the following discussion. 
\end{remark}

\section{Posterior conditions and bifurcation type}
\label{sec:posterior}

\subsection{Cumulant identities}
The posterior law of the clean datum given the noisy observation $X_\eta=x$ is
\begin{equation}\label{eq:posterior}
  \pi_{x,\eta}(dy):=\mathbb{P}(X_0\in dy\mid X_\eta=x)
  =\frac{\exp\bigl(-\|x-y\|^2/(2\eta)\bigr)\,p_0(dy)}{\int_{\mathbb{R}^d}\exp\bigl(-\|x-y'\|^2/(2\eta)\bigr)\,p_0(dy')} .
\end{equation}
Because of the Gaussian factor, $\pi_{x,\eta}$ has finite moments of every order for any probability measure $p_0$. 
Since here we deal with $\varphi_\eta$, the logarithm of $p_\eta$, 
we are rather interested in the \textit{cumulants}, that is
the derivatives of the logarithm of the moment-generating function.
They arise naturally because $p_\eta$ is itself, up to a Gaussian factor, a moment-generating function of the posterior, so that the derivatives of its logarithm $\varphi_\eta$ are posterior cumulants.
Namely, we define, for $Y \sim \pi_{x,\eta}$ 
\begin{equation}
\kappa_r:=\nabla_\theta^r\log\mathbb{E}_{x,\eta}[e^{\langle\theta,Y\rangle}]\big|_{\theta=0}
\label{eq:cumulants}
\end{equation}
Letting
\begin{equation}
m_\eta(x):=\mathbb{E}_{x,\eta}[Y],\qquad \zeta:=Y-m_\eta(x),\qquad C_\eta(x):=\mathbb{E}_{x,\eta}\bigl[\zeta\zeta^\top\bigr], 
\end{equation}
we have the identities 
\begin{equation}\label{eq:kappa}
\begin{aligned}
    &  \kappa_1=m_\eta,\qquad \kappa_2=C_\eta, \\[2mm]  & \kappa_3(u,v,w)=\mathbb{E}_{x,\eta}\bigl[\langle u,\zeta\rangle\langle v,\zeta\rangle\langle w,\zeta\rangle\bigr],
    \qquad 
\kappa_4(v,v,v,v)=\mathbb{E}_{x,\eta}\langle v,\zeta\rangle^4-3\,(v^\top C_\eta v)^2
\end{aligned}
\end{equation}
Remarkably, all cumulants of order $r\ge3$ vanish for a Gaussian law, so they measure how far the posterior is from Gaussian. 

\begin{lemma}[Cumulant representation]\label{lem:cumulants}
For every $\eta>0$ and $x\in\mathbb{R}^d$:
\begin{enumerate}
\item[(i)] the spatial jets of the log-density are
\begin{align}
  \nabla_x\varphi_\eta(x)&=\frac{m_\eta(x)-x}{\eta}, \label{eq:tweedie-grad}\\
  \nabla_x^2\varphi_\eta(x)&=\frac{C_\eta(x)-\eta I_d}{\eta^2}, \label{eq:tweedie-hess}\\
  \nabla_x^r\varphi_\eta(x)&=\frac{\kappa_r(x,\eta)}{\eta^r},\qquad r\ge3; \label{eq:jets}
\end{align}
\item[(ii)] the posterior cumulants satisfy, for every $r\ge1$ and $v\in\mathbb{R}^d$,
\begin{equation}\label{eq:dercum}
  D_x\kappa_r(x,\eta)[v]=\frac1\eta\,\kappa_{r+1}(x,\eta)(\,\cdot\,,\dots,\cdot\,,v).
\end{equation}
In particular, $D_xm_\eta=\eta^{-1}C_\eta$ and $D_xC_\eta[v]=\eta^{-1}\kappa_3(\,\cdot\,,\cdot\,,v)$.
\end{enumerate}
\end{lemma}

\begin{proof}
Completing the square in \eqref{eq:posterior}, the moment-generating function of the posterior is
\begin{equation}\label{eq:mgf-post}
\mathbb{E}_{x,\eta}\bigl[e^{\langle\theta,Y\rangle}\bigr]=\exp\bigl(\psi_\eta(x+\eta\theta)-\psi_\eta(x)\bigr),
  \qquad \psi_\eta(x):=\varphi_\eta(x)+\frac{\|x\|^2}{2\eta}.
\end{equation}
Taking the logarithm and differentiating $r$ times in $\theta$ at $\theta=0$, \eqref{eq:cumulants} gives $\kappa_r(x,\eta)=\eta^r\nabla_x^r\psi_\eta(x)$ for every $r\ge1$. Differentiating once more in $x$ gives (ii). Since $\nabla_x^r\psi_\eta$ differs from $\nabla_x^r\varphi_\eta$ only by $x/\eta$ for $r=1$ and by $I_d/\eta$ for $r=2$, (i) follows.
\end{proof}
Identities \eqref{eq:tweedie-grad}--\eqref{eq:jets} are classical: \eqref{eq:tweedie-grad} is Tweedie's formula~\cite{Robbins1956, Efron2011}, and the link between higher derivatives of $\log p_\eta$ and posterior cumulants is discussed in~\cite{Efron2011}, and for diffusion models in~\cite{Meng2021}.

\medskip

\noindent The reduced equations of~\Cref{sec:LS} also involve one derivative with respect to the noise level.
\begin{lemma}[Noise derivative]\label{lem:noise}
For every $\eta>0$:
\begin{enumerate}
\item[(i)] the log-density satisfies
\begin{equation}\label{eq:heatphi}
  \partial_\eta\varphi_\eta=\tfrac12\,\Delta_x\varphi_\eta+\tfrac12\,\|\nabla_x\varphi_\eta\|^2 
\end{equation}
\item[(ii)] if $x\in\Xi_\eta$ and $v\in\mathbb{R}^d$
then
\begin{equation}\label{eq:mixed-derivative}
  \partial_v\partial_\eta\varphi_\eta(x)=\frac{M_v(x,\eta)}{2\eta^3},\qquad
  M_v(x,\eta):=\mathbb{E}_{x,\eta}\bigl[\langle v,\zeta\rangle\,\|\zeta\|^2\bigr]=\sum_{j=1}^d\kappa_3(x,\eta)(v,e_j,e_j),
\end{equation}
for any orthonormal basis $(e_j)_{j=1}^d$ of $\mathbb{R}^d$.
\end{enumerate}
\end{lemma}

\begin{proof}
(i) $p_\eta$ solves the heat equation $\partial_\eta p_\eta=\tfrac12\Delta_xp_\eta$. Writing $p_\eta=e^{\varphi_\eta}$ gives $\Delta_xp_\eta=p_\eta\bigl(\Delta_x\varphi_\eta+\|\nabla_x\varphi_\eta\|^2\bigr)$, and dividing by $p_\eta$ yields \eqref{eq:heatphi}.
(ii) Differentiating \eqref{eq:heatphi} along $v$ gives $\partial_v\partial_\eta\varphi_\eta=\tfrac12\partial_v\Delta_x\varphi_\eta+\langle\nabla_x\varphi_\eta,\nabla_x^2\varphi_\eta\,v\rangle$, and the last term vanishes at $x\in\Xi_\eta$. By \eqref{eq:tweedie-hess}, $\Delta_x\varphi_\eta=\eta^{-2}\operatorname{tr}C_\eta-d/\eta$, and by \eqref{eq:dercum}, $\partial_v\Delta_x\varphi_\eta=\eta^{-2}\operatorname{tr}\bigl(D_xC_\eta[v]\bigr)=\eta^{-3}\sum_j\kappa_3(e_j,e_j,v)=\eta^{-3}M_v$.
\end{proof}

\Cref{lem:cumulants} implies the following results of immediate proof.  
\begin{corollary}[Critical set and spectrum]\label{cor:critical}
For every $\eta>0$ and $x\in\mathbb{R}^d$:
\begin{enumerate}
\item[(i)] $x\in\Xi_\eta$ if and only if $m_\eta(x)=x$
\item[(ii)] $H_\eta(x)$ and $C_\eta(x)$ have the same eigenvectors, and
\begin{equation}\label{eq:HCspec}
  C_\eta(x)\,v=\xi\,v\quad\Longrightarrow\quad H_\eta(x)\,v=\frac{\xi-\eta}{\eta^2}\,v .
\end{equation}
Since $C_\eta \succeq 0$ and $\xi \ge 0$, 
$H_\eta(x)\succeq-\eta^{-1}I_d$, and each eigenvalue of $H_\eta(x)$ is negative, zero or positive according to whether the corresponding eigenvalue of $C_\eta(x)$ is smaller, equal or larger than $\eta$
\item[(iii)] a point $x\in\Xi_\eta$ is a nondegenerate mode if and only if $\xi_{\max}(C_\eta(x))<\eta$, and $(x_c,\eta_c)$ is a soft critical point if and only if
\begin{equation}\label{eq:etac}
  \eta_c=\xi_{\max}\bigl(C_{\eta_c}(x_c)\bigr).
\end{equation}
In that case $E_c$ is also the eigenspace of $C_{\eta_c}(x_c)$ for $\eta_c$, and the transverse gap is $\gamma_c=(\eta_c-\xi_2)/\eta_c^2$, where $\xi_2<\eta_c$ is the largest remaining eigenvalue of $C_{\eta_c}(x_c)$
\item[(iv)] Taking $e_1 = v$ in \eqref{eq:mixed-derivative},
gives $M_v = \kappa_3(v,v,v) + \sum_{j\ge 2}\kappa_3(v,e_j,e_j)$. The first term measures the skewness of the posterior along $v$ only, whereas $M_v$ involves the posterior fluctuations in all directions. 
These two quantities will provide, respectively,
the quadratic coefficient and the time-transversality coefficient of the reduced fold equation.
\end{enumerate}
\end{corollary}

\begin{remark}[Empirical distribution]\label{rem:empirical}
For $p_0=\frac1N\sum_{i=1}^N\delta_{y_i}$ the posterior is the discrete law $\pi_{x,\eta}=\sum_iw_i\,\delta_{y_i}$, with softmax weights
\begin{equation}\label{eq:softmax}
  w_i(x,\eta)=\frac{\exp\bigl(-\|x-y_i\|^2/(2\eta)\bigr)}{\sum_{j=1}^N\exp\bigl(-\|x-y_j\|^2/(2\eta)\bigr)} .
\end{equation}
All the preceding quantities are therefore finite weighted sums. With $\zeta_i:=y_i-m_\eta(x)$ we have $m_\eta(x)=\sum_iw_iy_i$, and for $r=2,3$
\begin{equation}\label{eq:empirical-cum}
  \kappa_r(x,\eta)(u_1,\dots,u_r)=\sum_iw_i\prod_{k=1}^r\langle u_k,\zeta_i\rangle ,
\end{equation}
while $\kappa_4(v,v,v,v)=\sum_iw_i\langle v,\zeta_i\rangle^4-3\bigl(\sum_iw_i\langle v,\zeta_i\rangle^2\bigr)^2$ and $M_v=\sum_iw_i\langle v,\zeta_i\rangle\|\zeta_i\|^2$. 
\end{remark}
 
% one has explicitly
% \begin{align*}
% m_\eta(x)&=\textstyle\sum_i w_iy_i, &
% C_\eta(x)&=\textstyle\sum_i w_i\zeta_i\zeta_i^{\!\top},\\
% \kappa_3(x,\eta)(v,v,v)&=\textstyle\sum_i w_i\ip v{\zeta_i}^3, &
% \kappa_4(x,\eta)(v,v,v,v)&=\textstyle\sum_i w_i\ip v{\zeta_i}^4-3\,(v^{\!\top}Cv)^2,\\
% \kappa_3(x,\eta)(v,v,\cdot)&=\textstyle\sum_i w_i\ip v{\zeta_i}^2\zeta_i, &
% \textstyle\sum_j\kappa_3(x,\eta)(v,e_j,e_j)&=\textstyle\sum_i w_i\ip v{\zeta_i}|\zeta_i|^2 .
% \end{align*}

\subsection{Lyapunov--Schimdt reduction}
\label{sec:LS} 
From now on $(x_c,\eta_c)$ is a soft critical point in the sense of~\cref{def:soft}, and we set
\begin{equation}\label{eq:notation-c}
  H_c:=H_{\eta_c}(x_c),\qquad C_c:=C_{\eta_c}(x_c),\qquad \chi:=\eta-\eta_c .
\end{equation}
Posterior cumulants $\kappa_r$ are evaluated at $(x_c,\eta_c)$ unless stated otherwise. Since $\varphi_\eta(x)$ is $C^\infty$ in $(x,\eta)$ for every $p_0$ (\Cref{sec:noising}), all maps constructed below are $C^\infty$.

\begin{assumption}[Simple soft critical point]\label{ass:simple}
We assume that $\dim\ker H_c=1$; equivalently, by~\cref{cor:critical}, $\eta_c=\lambda_{\max}(C_c)$ is a simple eigenvalue of $C_c$.
\end{assumption}

\medskip
\noindent We will show in~\Cref{sec:critical-dimension} that simple soft critical points are the generic case if no further specific 
conditions are present. 

\smallskip

\noindent Before presenting the LS reduction, we need to introduce the 
spectral splitting operators which allow to obtain equations along the critical and transverse directions 
of the speciation.

\begin{lemma}[Spectral splitting]\label{lem:split}
Let $e_c\in\mathbb{R}^d$ be a unit vector with $\ker H_c=\operatorname{span}\{e_c\}$ and define
\begin{equation}\label{eq:projectors}
  P_c:=e_ce_c^\top,\qquad P_s:=I_d-P_c,\qquad E_s:=\operatorname{Ran}P_s=e_c^\perp .
\end{equation}
Then:
\begin{enumerate}
\item[(i)] $P_c$ and $P_s$ are complementary orthogonal projectors:
  $P_c=P_c^\top=P_c^2$, $P_s=P_s^\top=P_s^2$, $P_cP_s=0$;
\item[(ii)] since $H_c$ is symmetric, $E_s=(\ker H_c)^\perp=\operatorname{Ran}H_c$, and
  $\mathbb{R}^d=\ker H_c\oplus^{\perp}E_s$;
\item[(iii)] $H_cP_c=P_cH_c=0$ and $H_cP_s=P_sH_c=H_c$; hence $E_s$ is
  $H_c$-invariant and $H_c|_{E_s}:E_s\to E_s$ is invertible. By~\cref{ass:simple} it is negative definite, and by \cref{lem:cumulants}
  \begin{equation}\label{eq:Hsinv}
    \bigl(H_c|_{E_s}\bigr)^{-1}=\eta_c^2\,\bigl((C_c-\eta_cI_d)|_{E_s}\bigr)^{-1},
    \qquad
    \bigl\|(H_c|_{E_s})^{-1}\bigr\|=\gamma_c^{-1},
  \end{equation}
  where $\gamma_c:=\min\{|\lambda|:\lambda\in\operatorname{sp}(H_c)\setminus\{0\}\}>0$ is the
  transverse spectral gap.
\end{enumerate}
\end{lemma}
 
\begin{theorem}[Lyapunov--Schmidt reduction of the score]\label{thm:LS}
Under~\cref{ass:simple} there exist $\varepsilon,\delta>0$, a
neighbourhood $\mathcal U$ of $x_c$, and a $C^\infty$ map
\begin{equation}\label{eq:slaving}
  V_\eta:(-\varepsilon,\varepsilon)\times I_c\to E_s,\qquad
  I_c:=(\eta_c-\delta,\eta_c+\delta),\qquad
  V_{\eta_c}(0)=0,\quad \partial_uV_{\eta_c}(0)=0,
\end{equation}
such that, for $\eta\in I_c$ and $x\in\mathcal U$,
\begin{equation}\label{eq:LSequiv}
  S_\eta(x)=0
  \iff
  x=x_c+u\,e_c+V_\eta(u),\quad u=e_c^\top(x-x_c)\in(-\varepsilon,\varepsilon),
  \quad \Phi_\eta(u)=0,
\end{equation}
where the scalar bifurcation function is
\begin{equation}\label{eq:bif}
  \Phi_\eta(u):=e_c^\top S_\eta\bigl(x_c+u\,e_c+V_\eta(u)\bigr).
\end{equation}
\end{theorem}
 
\begin{proof}
Write $x=x_c+u\,e_c+v$ with $u\in\mathbb{R}$, $v\in E_s$. Since
$S_\eta=P_cS_\eta+P_sS_\eta$ and $P_cS_\eta=(e_c^\top S_\eta)\,e_c$, the equation
$S_\eta(x)=0$ is equivalent to the pair
\begin{align}
  \text{transverse:}&\quad P_sS_\eta(x_c+u\,e_c+v)=0, \label{eq:transv}\\
  \text{critical:}&\quad e_c^\top S_\eta(x_c+u\,e_c+v)=0. \label{eq:crit}
\end{align}
Let $G_\eta(u,v):=P_sS_\eta(x_c+u\,e_c+v)\in E_s$. Then $G_{\eta_c}(0,0)=0$ because
$x_c\in\Xi_{\eta_c}$, and, by~\cref{lem:split}(iii),
\[
  D_vG_{\eta_c}(0,0)=P_sH_c|_{E_s}=H_c|_{E_s}:E_s\to E_s
\]
is invertible. The IFT yields $\varepsilon,\delta>0$, a neighbourhood $\mathcal N$ of $0$ in $E_s$, and a unique $C^\infty$ map $V:(-\varepsilon,\varepsilon)\times I_c\to\mathcal N$ such that $v=V_\eta(u)$ is the only solution of \eqref{eq:transv} in $\mathcal N$, with $V_{\eta_c}(0)=0$. We take $\mathcal U:=\{x_c+u\,e_c+v:\ |u|<\varepsilon,\ v\in\mathcal N\}$. Differentiating the identity $G_{\eta}(u,V_\eta(u))\equiv0$ with respect to $u$ at $(0,\eta_c)$ gives
\begin{equation}
  H_c|_{E_s}\,\partial_uV(0,\eta_c)+P_sH_c\,e_c=0 .
\end{equation}
Since $H_ce_c=0$, invertibility of $H_c|_{E_s}$ gives $\partial_uV_{\eta_c}(0)=0$.
Substituting $v=V_\eta(u)$ into \eqref{eq:crit} gives \eqref{eq:LSequiv}--\eqref{eq:bif};
finally $u=e_c^\top(x-x_c)$ because $V_\eta(u)\in E_s=e_c^\perp$.
\end{proof}
 % Note that $V(0,\eta)$ need not vanish for $\eta\neq\eta_c$: in general no branch
% of critical points passes through $x_c$ (this happens only under additional
% structure, see Section~\ref{sec:pitchfork}).

\smallskip
 
\noindent 
It is useful for the following to introduce a reduced potential 
based on the LS reduction variables, which however shares the location of the critical points and their nature with $\varphi_t$ (and thus with $p_t$).

\begin{lemma}[Reduced potential; proof in~\Cref{app:lem:reduced}]\label{lem:reduced}
Let $x(u,\eta):=x_c+u\,e_c+V_\eta(u)$ and define the reduced log-density
\begin{equation}\label{eq:redpot}
  \widetilde U_\eta(u):=\varphi_\eta\bigl(x(u,\eta)\bigr),\qquad (u,\eta)\in(-\varepsilon,\varepsilon)\times I_c .
\end{equation}
Then $\widetilde U_\eta$ describes both the location and the type of the critical points of $p_\eta$ in $\mathcal U$:
\begin{enumerate}
\item[(i)] (location) $D_u\widetilde U_\eta(u)=\Phi_\eta(u)$, so the critical points of $p_\eta$ in $\mathcal U$ coincide with the points $x(u_*,\eta)$ with $\widetilde D_\eta U_\eta(u_*)=0$;
\item[(ii)] (type) possibly shrinking $\varepsilon$ and $\delta$, at every such point $H_\eta(x(u_*,\eta))$ is congruent to $\operatorname{diag}\bigl(\widetilde D_{uu}U_\eta(u_*),\,N\bigr)$ with $N$ negative definite. Hence $x(u_*,\eta)$ is a mode of $p_\eta$ if $\widetilde D_{uu}U_\eta(u_*)<0$ and a saddle with one unstable direction if $D_{uu}\widetilde U_\eta(u_*)>0$.
\end{enumerate}
\end{lemma}

\subsection{Classification of the bifurcation}\label{sec:classif}
We expand $\Phi_\eta(u)$ around $(u,\eta)=(0,\eta_c)$, with $\chi:=\eta-\eta_c$. At the soft critical point $\Phi_{\eta_c}(0)=0$, because $x_c\in\Xi_{\eta_c}$, and $\partial_u\Phi_{\eta_c}(0)=e_c^\top H_ce_c=0$, because $V_{\eta_c}'(0)=0$. Letting
\begin{equation}\label{eq:ABC}
A=\partial_u^2\Phi_{\eta_c}(0),\qquad B:=\partial_\eta\Phi_{\eta_c}(0),\qquad C:=\partial_u^3\Phi_{\eta_c}(0),\qquad D:=\partial_u\partial_\eta\Phi_{\eta_c}(0),
\end{equation}
we obtain
\begin{equation}\label{eq:Phi-joint}
  \Phi_\eta(u)=B\,\chi+D\,\chi\,u+\frac A2\,u^2+\frac C6\,u^3+O\bigl(\chi^2+|\chi|\,u^2+u^4\bigr)
\end{equation}
The dominating term in~\cref{eq:Phi-joint} depends on the relative size of $u$ and $\chi$: on the branches $u_\pm \approx \pm \sqrt{-2B\chi/A}=O(|\chi|^{1/2})$, so that $B\chi$ and $\frac A2u^2$ are of order $|\chi|$, while $D\chi u$ and $\frac C6u^3$ are only of order $|\chi|^{3/2}$. For $A\ne 0, B\ne 0$, the leading balance is thus $B\chi+\frac A2u^2$, which gives a fold. Under a reflection symmetry $\Phi_\eta$ is odd in $u$, so $A=B=0$, and the leading balance $D\chi u+\frac C6u^3$ gives a pitchfork. Since $\Phi_\eta=\widetilde U_\eta'$ by~\cref{lem:reduced}(i) we have also  
\begin{equation}\label{eq:ABC_1}
  A=\widetilde U^{'''}_{\eta_c}(0),\qquad C=\widetilde U^{'''}_{\eta_c}(0), \qquad 
  B=\partial_\eta\widetilde U_\eta'(0), \qquad 
  D=\partial_\eta\widetilde U_\eta''(0) 
\end{equation}

\begin{proposition}[Posterior form of the reduced coefficients; proof in~\Cref{app:prop:coeff}]\label{prop:coeff}
Under~\cref{ass:simple}, let all cumulants be evaluated at $(x_c,\eta_c)$ and set
\begin{equation}\label{eq:m3kc}
  m_3:=\mathbb{E}_{x_c,\eta_c}\bigl[\|\zeta\|^2\zeta\bigr]=\sum_{j=1}^d\kappa_3(e_j,e_j,\cdot\,),\quad
  k_c:=P_s\,\kappa_3(e_c,e_c,\cdot\,),\quad
  R_s:=\bigl((C_c-\eta_cI_d)|_{E_s}\bigr)^{-1},
\end{equation}
where $(e_j)_{j=1}^d$ is any orthonormal basis of $\mathbb{R}^d$. Then
\begin{align}
  A&=\frac{\kappa_3(e_c,e_c,e_c)}{\eta_c^3}, \label{eq:A}\\
  B&=\frac{\langle e_c,m_3\rangle}{2\eta_c^3}=\frac{M_{e_c}(x_c,\eta_c)}{2\eta_c^3}, \label{eq:B}\\
  C&=\frac{1}{\eta_c^4}\Bigl[\kappa_4(e_c,e_c,e_c,e_c)-3\,\langle k_c,R_s\,k_c\rangle\Bigr], \label{eq:C}\\
  D&=\frac{1}{2\eta_c^4}\Bigl[\sum_{j=1}^d\kappa_4(e_c,e_c,e_j,e_j)-\kappa_3\bigl(e_c,e_c,R_s\,P_sm_3\bigr)\Bigr]. \label{eq:D}
\end{align}
\end{proposition}

% =====================================================================
%  Sections 5.3.1 (generic fold) and 5.3.2 (symmetric pitchfork) -- revised.
%  Notation as in Sections 5.2-5.3: Phi_eta(u), V_eta(u), \widetilde U_eta(u),
%  x(u,eta) = x_c + u e_c + V_eta(u), chi = eta - eta_c, coefficients A, B, C, D
%  of (5.28) and their posterior form in Proposition 5.8.
%  \dot\lambda_c is replaced by D everywhere; old Proposition 5.13 (crossing rate)
%  is absorbed by Proposition 5.8 and replaced here by one paragraph.
%  Labels used: eq:Phi-joint (5.29), prop:coeff (Prop. 5.8), lem:reduced (Lemma 5.7),
%  thm:LS (Thm. 5.6), eq:transv (5.22), eq:posterior (5.1), ass:simple (Assumption 1).
% =====================================================================

\subsubsection{Generic fold}\label{sec:fold}

We consider the conditions
\begin{enumerate}
\item[(F1)] $A\neq0$, i.e.\ $\kappa_3(e_c,e_c,e_c)\neq0$ (quadratic nondegeneracy);
\item[(F2)] $B\neq0$, i.e.\ $M_{e_c}(x_c,\eta_c)\neq0$ (transversality in $\eta$).
\end{enumerate}

\begin{proposition}[Generic fold]\label{prop:fold}
Under (F1)--(F2), possibly shrinking $\varepsilon$ and $\delta$:
\begin{enumerate}
\item[(i)] if $AB\,\chi>0$, $p_\eta$ has no critical point in $\mathcal U$;
\item[(ii)] if $AB\,\chi<0$, $p_\eta$ has exactly two critical points in $\mathcal U$, with critical amplitudes
\begin{equation}\label{eq:foldbranches}
  u_\pm(\eta)=\pm\sqrt{-\frac{2B}{A}\,\chi}+O(|\chi|),
\end{equation}
one a mode and the other a saddle with one unstable direction.
\end{enumerate}
In particular $(x_c,\eta_c)$ is a density speciation: the number of local maxima in $\mathcal U$ changes from $0$ to $1$. The mode--saddle pair is created along the reverse process, i.e.\ it exists for $\eta<\eta_c$, if and only if $AB>0$.
\end{proposition}

\begin{proof}
By \eqref{eq:Phi-joint},
\begin{equation}\label{eq:foldred}
  \Phi_\eta(u)=B\,\chi+\frac A2\,u^2+O\bigl(|u|^3+|u|\,|\chi|+\chi^2\bigr).
\end{equation}
Since $\Phi_{\eta_c}'(0)=0$ and $\Phi_{\eta_c}''(0)=A\neq0$, the implicit function theorem applied to $\Phi_\eta'(u)=0$ gives a unique $u_m(\eta)=O(|\chi|)$ with $\Phi_\eta'(u_m(\eta))=0$, and, after shrinking $\varepsilon$ and $\delta$, $\Phi_\eta$ is strictly convex (if $A>0$) or strictly concave (if $A<0$) on $(-\varepsilon,\varepsilon)$. Hence $\Phi_\eta$ has two, one or no zeros according to whether $A\,\Phi_\eta(u_m(\eta))$ is negative, zero or positive. Since $\Phi_\eta(u_m(\eta))=B\chi+O(\chi^2)$, this gives (i) and the number of zeros in (ii). Solving \eqref{eq:foldred} for $u^2$ gives \eqref{eq:foldbranches}. Finally $\Phi_\eta'(u_\pm)=A\,u_\pm+O(|\chi|)$ has opposite signs on the two zeros, so by ~\cref{lem:reduced}(ii) one is a mode and the other a saddle with one unstable direction. The pair exists for $\chi<0$ if and only if $AB>0$.
\end{proof}

% A generic speciation thus implies the birth of a mode--saddle pair, not the splitting of an existing mode. No branch of critical points passes through $(x_c,\eta_c)$, so there is no eigenvalue crossing zero along a branch.

\begin{remark}[Barrier exponent of the fold]\label{rem:foldbarrier}
By~\cref{lem:reduced}(i), the log-density barrier between the mode and the saddle is $\Delta\varphi=|\widetilde U_\eta(u_+)-\widetilde U_\eta(u_-)|=\bigl|\int_{u_-}^{u_+}\Phi_\eta(u)\,du\bigr|$, and from \eqref{eq:foldred}
\begin{equation}\label{eq:foldbarrier}
  \Delta\varphi=\frac{4\sqrt2}{3}\,\frac{|B|^{3/2}}{|A|^{1/2}}\,|\eta-\eta_c|^{3/2}\bigl(1+O(|\eta-\eta_c|^{1/2})\bigr).
\end{equation}
The exponent $3/2$ in $\eta$ is a characteristic signature of the fold.
\end{remark}

% ---------------------------------------------------------------------
\subsubsection{Symmetric pitchfork}\label{sec:pitchfork}
We now assume the following symmetry:
\begin{enumerate}
\item[(P0)] there exists an orthogonal reflection $R$ with push--forward $R_\#p_0=p_0$ and mirror relations $Rx_c=x_c$ and $Re_c=-e_c$.
\end{enumerate}

\begin{lemma}[Consequences of the symmetry]\label{lem:sym}
Under (P0), for every $\eta\in I_c$, taking $\mathcal U$ to be $R$-invariant:
\begin{enumerate}
\item[(i)] $\varphi_\eta\circ R=\varphi_\eta$, $S_\eta(Rx)=RS_\eta(x)$, $H_\eta(Rx)=RH_\eta(x)R$, and $RE_s=E_s$;
\item[(ii)] $V_\eta(-u)=RV_\eta(u)$; hence $\Phi_\eta$ is odd and $\widetilde U_\eta$ is even. In particular $A=B=0$, that is, $\kappa_3(e_c,e_c,e_c)=0$ and $M_{e_c}=0$;
\item[(iii)] $x_0(\eta):=x_c+V_\eta(0)$ is a branch of critical points of $p_\eta$ with $x_0(\eta_c)=x_c$, contained in $\operatorname{Fix}(R)$.
\end{enumerate}
\end{lemma}

\begin{proof}
(i) The Gaussian kernel in \eqref{eq:posterior} is invariant under orthogonal maps, so $p_\eta\circ R=p_\eta$, and the identities for $S_\eta$ and $H_\eta$ follow by differentiation. Since $R$ is orthogonal and $Re_c=-e_c$, it preserves $e_c^\perp=E_s$.
(ii) $P_s$ commutes with $R$. Applying $P_s$ to $S_\eta(R\,x(u,\eta))=RS_\eta(x(u,\eta))$ shows that $R\,x(u,\eta)=x_c-u\,e_c+RV_\eta(u)$ solves \eqref{eq:transv} with amplitude $-u$. By uniqueness in ~\cref{thm:LS}, $V_\eta(-u)=RV_\eta(u)$. Then $\Phi_\eta(-u)=e_c^\top S_\eta(Rx(u,\eta))=(Re_c)^\top S_\eta(x(u,\eta))=-\Phi_\eta(u)$ and $\widetilde U_\eta(-u)=\varphi_\eta(Rx(u,\eta))=\widetilde U_\eta(u)$. Since $\Phi_\eta$ is odd, its even derivatives at $u=0$ vanish for every $\eta$, so $A=B=0$.
(iii) $\Phi_\eta(0)=0$ for every $\eta$, and $V_\eta(0)=RV_\eta(0)$.
\end{proof}

Along the symmetric branch, $\Phi_\eta'(0)=\widetilde U_\eta''(0)$ vanishes exactly when $H_\eta(x_0(\eta))$ is singular (\cref{lem:reduced}(ii)). Its derivative at $\eta_c$ is the coefficient $D$ of \eqref{eq:Phi-joint}. By first-order perturbation theory, since $H_ce_c=0$, $D=\lambda_c'(\eta_c)$ is the crossing rate of the continuation $\lambda_c(\eta)$ of the zero eigenvalue of $H_\eta(x_0(\eta))$. Its posterior form is given in~\cref{prop:coeff}. Under (P0), the velocity $\dot x$ of the branch appearing in its proof is $x_0'(\eta_c)$, and $\dot x=0$ when the symmetric point does not move, e.g.\ for $x_c=0$ and $R=-I_d$; in that case only the first term of \eqref{eq:D} survives.

\begin{proposition}[Recognition of the pitchfork {\cite[Ch.~VI]{GolubitskySchaeffer1985}}]\label{prop:pitchfork}
Assume (P0) and
\begin{enumerate}
\item[(P1)] $D\neq0$ (transversal crossing along the symmetric branch);
\item[(P2)] $C\neq0$ (cubic nondegeneracy).
\end{enumerate}
Then $\Phi_\eta(u)$ is $\mathbb Z_2$-equivalent near $(0,\eta_c)$ to the normal form $\epsilon\,u^3+\varsigma\,\chi\,u$, with $\epsilon=\operatorname{sgn}C$ and $\varsigma=\operatorname{sgn}D$. Explicitly,
\begin{equation}\label{eq:pitchred}
  \Phi_\eta(u)=u\Bigl[D\,\chi+\frac C6\,u^2+O\bigl(u^4+|\chi|\,u^2+\chi^2\bigr)\Bigr],
\end{equation}
and the nontrivial zeros satisfy
\begin{equation}\label{eq:pitchbranches}
  u_\pm^2=-\frac{6D}{C}\,\chi+O(\chi^2),
\end{equation}
so that they exist exactly on the side $D\chi/C<0$.
\end{proposition}

\begin{proof}
Since $\Phi_\eta$ is odd and smooth, $\Phi_\eta(u)=u\,g(u^2,\eta)$ with $g$ smooth, and \eqref{eq:Phi-joint} with $A=B=0$ gives \eqref{eq:pitchred}. Setting the bracket to zero gives \eqref{eq:pitchbranches}. The $\mathbb Z_2$-equivalence is the recognition theorem of~\cite[Ch.~VI]{GolubitskySchaeffer1985}.
\end{proof}

On the symmetric branch $\Phi_\eta'(0)=D\chi+O(\chi^2)$, and on the new branches $\Phi_\eta'(u_\pm)=-2D\chi+O(\chi^2)$ which has the sign of $C$. By~\cref{lem:reduced}(ii), no further condition on the transverse directions is needed to classify the critical points: if $C<0$ the new branches are modes, and if $C>0$ they are saddles with one unstable direction. The branches appear as the noise decreases (supercritical in the direction of generation) if and only if $D$ and $C$ have the same sign. The case relevant for the splitting of a mode into two is therefore $C<0$, $D<0$: for $\eta<\eta_c$ the symmetric point becomes a saddle and two symmetry-related modes emerge, so that $(x_c,\eta_c)$ is a density speciation ($1\to2$ maxima).

\begin{remark}[Barrier exponent of the pitchfork]\label{rem:pitchbarrier}
By~\cref{lem:reduced}(i) and \eqref{eq:pitchred}, $\widetilde U_\eta(u)-\widetilde U_\eta(0)=\tfrac12D\chi\,u^2+\tfrac1{24}C\,u^4+\cdots$, and for $C<0$ the barrier between the modes and the saddle is
\begin{equation}\label{eq:pitchbarrier}
  \Delta\varphi=\frac32\,\frac{D^2}{|C|}\,(\eta-\eta_c)^2\bigl(1+O(|\eta-\eta_c|)\bigr).
\end{equation}
The exponent $2$ is a characteristic signature of the pitchfork, to be compared 
against $3/2$ of the fold \eqref{eq:foldbarrier}.
A direct experimental test can detect which of the two 
singularities is then present. Notice that, since $\eta=\sigma^2$, $\eta-\eta_c=2\sigma_c(\sigma-\sigma_c)+O((\sigma-\sigma_c)^2)$, so the same exponents hold in $\sigma$.
\end{remark}

\begin{example}[Symmetric doublet (revisited)]\label{ex:doublet}
Let $p_0=\tfrac12\delta_{m}(e)+\tfrac12\delta_{-m}(e)$, with $m>0$ and $\|e\|=1$, and take $R=-I_d$, $x_c=0$. At $x=0$ the posterior is $\pm me$ with probability $\tfrac12$, so $m_\eta(0)=0$, i.e.\ $0\in\Xi_\eta$ for every $\eta$, and $C_\eta(0)=m^2ee^\top$, whence $H_\eta(0)=(m^2ee^\top-\eta I_d)/\eta^2$. The kernel appears at $\eta_c=m^2$ (i.e.\ $\sigma_c=m$), with $e_c=e$, $E_s=e^\perp$ and $H_c|_{E_s}=-m^{-2}I$. The posterior at $0$ is symmetric, so all odd cumulants vanish: $k_c=0$, $m_3=0$, and $x_0(\eta)\equiv0$. Moreover $\langle e,\zeta\rangle=\pm m$, so $\kappa_4(e,e,e,e)=m^4-3m^4=-2m^4$, while $\kappa_4(e,e,e_j,e_j)=0$ for $e_j\perp e$. By ~\cref{prop:coeff},
\begin{equation}
  C=-\frac{2}{m^4}<0,\qquad
  D=-\frac{1}{m^4}=\frac{d}{d\eta}\Bigl(\frac{m^2}{\eta^2}-\frac1\eta\Bigr)\Big|_{\eta=m^2}
\end{equation}
Both are negative: the pitchfork is supercritical in the direction of generation, the new branches are modes, and by \eqref{eq:pitchbranches} $u_\pm^2=3(\eta_c-\eta)+O(\chi^2)$. As an exact check, along $e$ the critical points solve $s=m\tanh(ms/\eta)$, and expanding $\tanh z=z-z^3/3+O(z^5)$ gives $s^2=3\eta^2(m^2-\eta)/m^4+O\bigl((m^2-\eta)^2\bigr)=3(\eta_c-\eta)+O(\chi^2)$.
\end{example}

\begin{remark}[Doublet of Gaussian clouds]\label{rem:gaussdoublet}
Let $p_0=\tfrac12\mathcal N(\mu,\Sigma)+\tfrac12\mathcal N(-\mu,\Sigma)$, with $\mu\neq0$ and $\Sigma\succeq0$. At $x=0$ the component posteriors have covariance $\eta\Sigma(\Sigma+\eta I_d)^{-1}$ and means $\pm\eta\,b$, with $b:=(\Sigma+\eta I_d)^{-1}\mu$, so that
\begin{equation}\label{eq:gaussdoublet-C}
  C_\eta(0)-\eta I_d=\eta^2\bigl(bb^\top-(\Sigma+\eta I_d)^{-1}\bigr).
\end{equation}
This is a rank-one perturbation of a negative definite matrix, so at most one eigenvalue can reach zero. By the matrix determinant lemma, $H_\eta(0)$ is singular if and only if
\begin{equation}\label{eq:gaussdoublet}
  \mu^\top(\Sigma+\eta_cI_d)^{-1}\mu=1,
\end{equation}
and then $\ker H_{\eta_c}(0)=\operatorname{span}\{\mu\}$. Since $\eta\mapsto\mu^\top(\Sigma+\eta I_d)^{-1}\mu$ decreases strictly from $\mu^\top\Sigma^{-1}\mu$ to $0$, the origin is a soft critical point for exactly one $\eta_c$ if $\mu^\top\Sigma^{-1}\mu>1$ (with the convention $\mu^\top\Sigma^{-1}\mu=+\infty$ if $\mu\notin\operatorname{Ran}\Sigma$), and never otherwise. The critical direction is always $e_c=\mu/\|\mu\|$: the anisotropy of $\Sigma$ shifts $\eta_c$ but does not rotate the splitting direction. If $\mu=m\,e$ with $\Sigma e=c\,e$, then $\eta_c=m^2-c$, and no speciation occurs at the origin when $m^2\le c$. Since (P0) holds with $R=-I_d$, the bifurcation falls in the pitchfork class.
\end{remark}

%%%%%%%%
\section{Dimension of the critical subspace}
\label{sec:critical-dimension}

We address here the critical point why a one-dimensional critical subspace is the generic situation, and when higher-dimensional kernels can occur. 
Throughout, we set
\begin{equation}\label{eq:Mdef}
  M_\eta(x):=C_\eta(x)-\eta I_d \in \operatorname{Sym}(\mathbb R^{d\times d}),
\end{equation}
so that $H_\eta(x)=M_\eta(x,\eta)/\eta^2$ by \eqref{eq:tweedie-hess}, and $\ker H_\eta(x)=\ker M_\eta(x)$ (\cref{cor:critical}).

\subsection{A counting argument}
Let $M:=M_\eta(x)$ have $\dim\ker M=q$. Every $M'\in\operatorname{Sym}(\mathbb{R}^d)$ close to $M$ can be written, with respect to the fixed splitting $\ker M\oplus(\ker M)^\perp$, as
\begin{equation}\label{eq:schur}
  M'=\begin{pmatrix}X&Y\\Y^\top&Z\end{pmatrix},\qquad
  X\in\operatorname{Sym}(\mathbb{R}^q),\ Y\in\mathbb{R}^{q\times(d-q)},\ Z\in\operatorname{Sym}(\mathbb{R}^{d-q})\ \text{invertible}.
\end{equation}
Since $M'$ is congruent to $\operatorname{diag}(L',Z)$, with Schur complement $L':=X-YZ^{-1}Y^\top\in\operatorname{Sym}(\mathbb{R}^q)$, we have $\dim\ker M'=\dim\ker L'$. Hence $\dim\ker M'=q$ if and only if $L'=0$. The map $M'\mapsto L'$ is a submersion, since its derivative with respect to $X$ is the identity, so near $M$ the matrices of corank $q$ form a smooth submanifold of $\operatorname{Sym}(\mathbb{R}^d)$ of codimension $q(q+1)/2$.

\smallskip

\noindent For $(x,\eta)$ to be a critical point of corank $q$ we have thus $d+q(q+1)/2$ scalar equations in the $d+1$ unknowns $(x,\eta)$. If the map $(x,\eta)\mapsto\bigl(S_\eta(x),M_\eta(x)\bigr)$ is transverse to $\{0\}\times\{\text{corank }q\}$, the regular value theorem shows that the soft critical points of corank $q$ form a set of dimension
\begin{equation}\label{eq:count}
  (d+1)-\Bigl(d+\frac{q(q+1)}{2}\Bigr)=1-\frac{q(q+1)}{2}.
\end{equation}
This set consists of isolated points for $q=1$ and is empty for $q\ge2$. Equivalently, the critical points of $p_\eta$ form, generically, curves in $(x,\eta)$-space, and a curve meets a set of codimension $q(q+1)/2$ only if $q(q+1)/2\le1$, that is, $q=1$. For $q=2$ the three conditions imply that two eigenvalues of $C_\eta(x)$ must coincide, which costs two conditions for real symmetric matrices (the von Neumann--Wigner non-crossing rule), and their common value must equal $\eta$, which costs one more.

\smallskip

\noindent Notice that transversality is a genericity assumption on $p_0$, not a consequence of the setting. It fails when the data have additional structure, typically a symmetry that forces several eigenvalues of $C_\eta(x)$ to coincide for every $\eta$. This is the object of the following example.

\medskip

\paragraph{A rotationally symmetric example with a two-dimensional critical subspace ($q=2$)}
Let $p_0$ be the uniform probability measure on the circle of radius $r$
in $\mathbb{R}^2$. 
A point on the circle is $
Y=r(\cos\theta,\sin\theta),
\, \theta\sim\mathrm{Unif}[0,2\pi)$.
The diffusion produces 
$X_\eta=Y+\sqrt{\eta}\,Z,
\, Z\sim\mathcal N(0,I_2)$.
Observing \(X_\eta=x\), the posterior of \(Y\) is proportional to $\exp\!\left(-\frac{\|x-Y\|^2}{2\eta}\right)$.
At $x=0$, the likelihood
is constant on the circle, the posterior $\pi_{0,\eta}$ is also
uniform on the circle, and
\begin{equation}
m_\eta(0)=0,
\qquad
C_\eta(0)
=\mathbb{E}_{0,\eta}[YY^\top]
=\frac{r^2}{2}I_2.
\end{equation}
The posterior identities of~\cref{lem:cumulants} therefore give
\begin{equation}
S_\eta(0)=0,
\qquad
H_\eta(0)
=\frac{C_\eta(0)-\eta I_2}{\eta^2}
=\frac{r^2/2-\eta}{\eta^2}I_2.
\end{equation}
At
$\eta_c=\frac{r^2}{2}$,
both eigenvalues of $H_c$ vanish. Thus
\begin{equation}
q=\dim\ker H_c=2.
\end{equation}
This simultaneous degeneracy is enforced by rotational symmetry. The reduced problem is therefore
genuinely two-dimensional and it cannot be
described by a scalar pitchfork normal form. Namely, writing $\rho=\|x\|$ and denoting by $\mathcal I_0$ the modified
Bessel function of the first kind, the smoothed density is radial and
satisfies
\begin{equation}
p_\eta(x)\propto
\exp\!\left(-\frac{\rho^2+r^2}{2\eta}\right)
\mathcal I_0\!\left(\frac{r\rho}{\eta}\right).
\end{equation}
Near $\rho=0$,
\begin{equation}
\log p_\eta(\rho)
=
\mathrm{const}
+\frac{r^2-2\eta}{4\eta^2}\rho^2
-\frac{r^4}{64\eta^4}\rho^4
+O(\rho^6).
\end{equation}
At $\eta=\eta_c$, the quadratic term vanishes while the quartic term is
negative. As $\eta$ decreases below $\eta_c$, the origin loses its
maximum and a \textit{circle of maxima} emerges, as required by rotational
symmetry.

\subsection{A priori bounds for mixtures}
\label{subsec:mixture-bound}
In this section we derive a more rigorous argument under the hypothesis that the data distribution is a finite mixture. This is also the framework of our other research work~\cite{softmodes2026}.
Let 
\begin{equation}
    p_0=\sum_{j=1}^K\alpha_jp_j,
    \qquad
    \alpha_j>0,
    \qquad
    \sum_{j=1}^K\alpha_j=1,
    \label{eq:data-mixture}
\end{equation}
where the \(p_j\) are the component probability measures. For each component, let
\begin{equation}
    Z_j(x,\eta)=
    \int_{\mathbb R^d}
    \exp\!\left(-\frac{\lVert x-y\rVert^2}{2\eta}\right)p_j(dy),
    \qquad
    w_j(x,\eta)
    =
    \frac{\alpha_jZ_j(x,\eta)}
         {\sum_{\ell=1}^K\alpha_\ell Z_\ell(x,\eta)}.
    \label{eq:posterior-component-weights}
\end{equation}
The full posterior is then a mixture, with weights \(w_j\), of the
componentwise posteriors
\[
    \pi^{(j)}_{x,\eta}(dy)
    \propto
    \exp\!\left(-\frac{\lVert x-y\rVert^2}{2\eta}\right)p_j(dy).
\]
Denote their means and covariances by \(m_j(x,\eta)\) and
\(C_j(x,\eta)\), respectively.

\begin{proposition}[Bound on the number of critical directions]
\label{prop:mixture-critical-bound}
The posterior covariance admits the decomposition
\begin{equation}
    C_\eta(x)=W_\eta(x)+B_\eta(x),
    \label{eq:total-covariance-decomposition}
\end{equation}
where
\begin{equation}
    W_\eta(x)
    =\sum_{j=1}^K w_jC_j,
    \qquad
    B_\eta(x)
    =\sum_{j=1}^K
      w_j(m_j-m_\eta)(m_j-m_\eta)^\top,
    \label{eq:within-between-covariance}
\end{equation}
and
\begin{equation}
    \operatorname{rank}B_\eta(x)\leq K-1.
    \label{eq:between-rank}
\end{equation}
If
\begin{equation}
    C_j(x,\eta)\prec\eta I_d
    \qquad
    \text{for every }j,
    \label{eq:subcritical-components}
\end{equation}
then \(C_\eta(x)\) has at most \(K-1\) eigenvalues greater than or equal to
\(\eta\). In particular,
\begin{equation}
    \dim\ker H_\eta(x)\leq K-1,
    \qquad
    \#\{\lambda>0:\lambda\in\sigma(H_\eta(x))\}\leq K-1.
    \label{eq:hessian-index-bound}
\end{equation}
\end{proposition}

\begin{proof}
Equation~\eqref{eq:total-covariance-decomposition} is the law of total
covariance: \(W_\eta\) is the posterior average of the covariance within each
component, while \(B_\eta\) is the covariance of the component posterior
means. Since
\[
    \sum_{j=1}^Kw_j(m_j-m_\eta)=0,
\]
the vectors \(m_j-m_\eta\) span a space of dimension at most \(K-1\), proving
\eqref{eq:between-rank}.
Condition~\eqref{eq:subcritical-components} implies
\[
    W_\eta(x)\prec\eta I_d,
    \qquad
    \lambda_1(W_\eta)<\eta.
\]
If $K>d$, the bound is trivial, since $C_\eta(x)$ has only $d$ eigenvalues; the statement is informative only for $K\le d$, which we assume from now on.
Order the eigenvalues nonincreasingly. Since
\(B_\eta\) is positive semidefinite and has rank at most \(K-1\),
\(\lambda_K(B_\eta)=0\). Weyl's inequality gives
\[
    \lambda_K(C_\eta)
    \leq
    \lambda_1(W_\eta)+\lambda_K(B_\eta)
    =\lambda_1(W_\eta)
    <\eta.
\]
Thus at most \(K-1\) eigenvalues of \(C_\eta\) can be greater than or equal to
\(\eta\). The conclusions for \(H_\eta\) follow from
\cref{lem:cumulants}.
\end{proof}

The decomposition \eqref{eq:total-covariance-decomposition} has a direct
interpretation. The within-component term \(W_\eta\) measures uncertainty
inside the individual clusters, whereas \(B_\eta\) measures the separation of
their posterior means. Under \eqref{eq:subcritical-components}, no individual
component can create a critical direction by itself; all criticality is caused
by competition between different components. Since \(K\) posterior means have
at most \(K-1\) independent relative displacements, at most \(K-1\) critical
directions can emerge.

\begin{remark}[Gaussian mixtures]
A very relevant case in which  condition~\eqref{eq:subcritical-components} holds for every
\(x\in\mathbb R^d\) and every \(\eta>0\)  is when the components of the mixture are Gaussians, $p_j=\mathcal N(\mu_j,\Sigma_j)$, including the limiting case of a point
    mass. The posterior covariance is then
    \begin{equation}
        C_j
        =\eta\Sigma_j(\Sigma_j+\eta I_d)^{-1}.
        \label{eq:gaussian-posterior-covariance}
    \end{equation}
An eigenvalue \(s\geq0\) of \(\Sigma_j\) produces the
posterior-covariance eigenvalue
 $\frac{\eta s}{s+\eta}<\eta$.
This remains valid when \(s=0\), and hence also for singular covariances and
point masses.
In particular,  for a mixture of two Gaussian 
    $\dim\ker H_\eta(x)\leq1$
at every point and every noise level. Consequently, every degenerate critical
point of such a mixture has a one-dimensional critical subspace.
\end{remark}

\begin{remark}[Effective number of critical components]
The bound in~\cref{prop:mixture-critical-bound} is governed in
practice by the components whose posterior weights \(w_j(x,\eta)\) are not
negligible. Negligible components may be grouped into a residual component,
provided that their contribution to the within-component term does not raise
\(\lambda_1(W_\eta)\) to the threshold \(\eta\). Thus, near a decision region
between two dominant clusters, the local covariance is effectively controlled
by two components even when the full dataset contains many classes. This gives
a structural explanation for the frequently observed sequence of approximately
binary splitting events (see also the discussion in~\cite[Sect.6]{softmodes2026}). 
\end{remark}

\section{Stability of equilibria under perturbations of the score}
\label{sec:stability}
Having established the above framework, from an applicative viewpoint it is interesting to study what happens when we add a perturbation on the model (see \cref{tab:dict} for a comparison with the mechanical model). Namely, we want to study the effect of perturbations on the score, typically occurring by learning it in an approximated manner via
a neural network or by applying a guidance.

\smallskip

\noindent We address separately four situations, in the order in which they are met along generation: 1) equilibria away from  speciations, 2) equilibria close to a speciation, 3) the speciation itself, and 4) equilibria after the last speciation.

\smallskip
 
\paragraph{1) Equilibria away from speciations}
Away from speciations the Hessian is uniformly invertible along a branch, and the equilibria depend Lipschitz-continuously on the perturbation.
 
\begin{proposition}[Uniform robustness]\label{prop:robust}
Let $x^\ast:[\eta_a,\eta_b]\to\mathbb{R}^d$ be a $C^1$ branch of critical points, $S_\eta(x^\ast(\eta))=0$, with $\|H_\eta(x^\ast(\eta))^{-1}\|\le\gamma_0^{-1}$ for all $\eta\in[\eta_a,\eta_b]$. For $|\varepsilon|$ small enough there is a unique branch $x^\ast_\varepsilon$ of zeros of the perturbed field near $x^\ast$, and
\begin{equation}\label{eq:robust}
  \sup_{\eta\in[\eta_a,\eta_b]}\|x^\ast_\varepsilon(\eta)-x^\ast(\eta)\|
  \le\frac{|\varepsilon|}{\gamma_0}\sup_{\eta\in[\eta_a,\eta_b]}\|\delta S_\eta(x^\ast(\eta))\|+O(\varepsilon^2).
\end{equation}
If $x^\ast$ is a branch of modes, the zeros $x^\ast_\varepsilon$ are stable, i.e.\ all eigenvalues of the perturbed Jacobian have negative real part.
\end{proposition}
 
\begin{proof}
Apply the implicit function theorem to $(x,\varepsilon)\mapsto S_\eta(x)+\varepsilon\,\delta S_\eta(x)$ at each $\eta$; compactness of $[\eta_a,\eta_b]$ makes the range of $\varepsilon$ uniform. To first order, $\delta x^\ast=-\varepsilon\,H_\eta(x^\ast)^{-1}\delta S_\eta(x^\ast)$, which gives \eqref{eq:robust}. Stability persists by continuity of the eigenvalues.
\end{proof}

\paragraph{2) Equilibria near a speciation: critical read--out}
Near a speciation $\gamma_0\to0$ and \eqref{eq:robust} degenerates. The next lemma shows that the degeneration is confined to the critical amplitude. Let $x^\star=x(u^\star,\eta)$ be a nondegenerate critical point in the neighbourhood $\mathcal U$ of~\cref{thm:LS}, i.e.\ $\Phi_\eta'(u^\star)\neq0$, and let $w^\star:=e_c+V_\eta'(u^\star)$ as in~\cref{lem:reduced}. We write the first-order response to a perturbation $\delta S$ as
\begin{equation}
  \delta x^\star=\delta u^\star\,w^\star+\delta z^\star,\qquad \delta z^\star\in E_s
\end{equation}
since $e_c^\top w^\star=1$ and $\delta u^\star=e_c^\top\delta x^\star$ is the change of the critical amplitude.
 
\begin{lemma}[Critical read-out]\label{lem:sens}
Up to $O(\|\delta S\|^2)$:
\begin{enumerate}
\item[(i)] (read-out) the error enters the critical equation through the scalar
\begin{equation}\label{eq:readout}
  h^\star:=w^{\star\top}\delta S(x^\star)=e_c^\top\delta S(x^\star)+V_\eta'(u^\star)^\top P_s\,\delta S(x^\star),
\end{equation}
where $V_\eta'(u^\star)=O(|u^\star|+|\chi|)$;
\item[(ii)] (critical amplitude) $\displaystyle\delta u^\star=-\frac{h^\star}{\Phi_\eta'(u^\star)}$;
\item[(iii)] (transverse component) $\delta z^\star=-W^{-1}P_s\,\delta S(x^\star)$, with $W:=P_sH_\eta(x^\star)|_{E_s}$, so that $\|\delta z^\star\|\le\gamma^{-1}\|P_s\delta S(x^\star)\|$, where $\gamma$ is the smallest modulus of the eigenvalues of $W$; $\gamma$ is close to the transverse gap $\gamma_c$ near $(x_c,\eta_c)$.
\end{enumerate}
\end{lemma}
 
\begin{proof}
Linearizing $S_\eta(x)+\delta S(x)=0$ at $x^\star$ gives $H_\eta(x^\star)\,\delta x^\star+\delta S(x^\star)=0$. By~\cref{lem:reduced}(ii), $H_\eta w^\star=\Phi_\eta'(u^\star)\,e_c$ and $w^{\star\top}H_\eta z=0$ for $z\in E_s$. Multiplying by $w^{\star\top}$ gives $\Phi_\eta'(u^\star)\,\delta u^\star+w^{\star\top}\delta S(x^\star)=0$, which is (i)--(ii). Applying $P_s$ and using $P_sH_\eta w^\star=0$ gives $W\,\delta z^\star+P_s\delta S(x^\star)=0$, which is (iii); $W$ is invertible by~\cref{lem:reduced}(ii). Finally, $V_\eta'(u^\star)=O(|u^\star|+|\chi|)$ because $V_{\eta_c}'(0)=0$ by~\cref{thm:LS}.
\end{proof}
 
\begin{remark}\label{rem:sens}
The transverse part $P_s\delta S$ of the error is absorbed by the spectral gap in (iii), and reaches the critical amplitude only through the tilt $V_\eta'$ of the slaving manifold, which vanishes at the speciation point. At first order, the score error is therefore read through the single scalar $e_c^\top\delta S$, however large its transverse components. The only possible amplification is in (ii): near a fold $\Phi_\eta'(u_\pm)=\pm\operatorname{sgn}(A)\sqrt{2|AB|\,|\chi|}+O(|\chi|)$, so that $|\delta u^\star|\sim|\chi|^{-1/2}$. The linear response breaks down at the fold itself, where it is replaced by the displacement of the fold described in the next paragraph.
\end{remark}

\medskip
 
\noindent \paragraph{3) Equilibria at  speciation}
At the speciation, the relevant point is to quantify how the speciation itself moves, in noise level and in space. 

\begin{proposition}[Displacement of a nondegenerate fold; proof in~\Cref{app:prop:foldpersist}]\label{prop:foldpersist}
Let $(x_c,\eta_c)$ satisfy (F1)--(F2) of~\Cref{sec:fold}, and set
\[
  h:=e_c^\top\delta S_{\eta_c}(x_c),\qquad
  h':=e_c^\top D_x\delta S_{\eta_c}(x_c)\,e_c,\qquad
  \delta z:=-(H_c|_{E_s})^{-1}P_s\,\delta S_{\eta_c}(x_c)\in E_s,
\]
\[
  \dot x:=-(H_c|_{E_s})^{-1}P_s\,\partial_\eta S_\eta(x_c)\big|_{\eta_c}\in E_s .
\]
There exists $\varepsilon_0>0$ such that, for $|\varepsilon|<\varepsilon_0$, the field $S_\eta+\varepsilon\,\delta S_\eta$ has a unique nondegenerate fold $(x_c(\varepsilon),\eta_c(\varepsilon))$ near $(x_c,\eta_c)$, depending $C^1$ on $\varepsilon$, with
\begin{align}
  \eta_c(\varepsilon)-\eta_c&=-\frac{h}{B}\,\varepsilon+O(\varepsilon^2), \label{eq:foldshift}\\
  x_c(\varepsilon)-x_c&=\varepsilon\bigl(u_c'\,e_c-\tfrac hB\,\dot x+\delta z\bigr)+O(\varepsilon^2),\qquad
  u_c'=\frac1A\Bigl(\frac DB\,h-h'-\frac{\langle k_c,\delta z\rangle}{\eta_c^3}\Bigr). \label{eq:foldshift-x}
\end{align}
\end{proposition}

Both the noise level and the critical amplitude of a generic speciation thus respond linearly to the read-out $h$, with gain $1/|B|$ for the noise level. The transverse part of the error enters only through $\delta z$, which is controlled by the spectral gap. The gain $1/|B|$ can be large: for the Trouser/Sneaker pair of~\Cref{sec:numerical} it is about $264$, so that a score error of $1\%$ along $e_c$ shifts $\eta_c$ by up to $10\%$.

\smallskip
 
\noindent Near a symmetric configuration the response is no longer linear. When the unperturbed speciation is a pitchfork, $A=B=0$ and \eqref{eq:foldshift} is meaningless; a symmetry-breaking error unfolds the pitchfork into a continuous branch and a fold, whose position scales differently.
 
\begin{proposition}[Unfolding of a pitchfork: cusp law; proof in~\Cref{app:prop:cusp}]\label{prop:cusp}
Assume (P0)--(P2) of~\Cref{sec:pitchfork}, and let $h:=e_c^\top\delta S_{\eta_c}(x_c)\neq0$. For $\varepsilon\neq0$ small enough, the perturbed field has a unique nondegenerate fold near $(x_c,\eta_c)$, at
\begin{equation}\label{eq:cusp}
  \bigl(\eta_c(\varepsilon)-\eta_c\bigr)^3=-\frac{9\,C}{8\,D^3}\,h^2\varepsilon^2\,\bigl(1+O(|\varepsilon|^{1/3})\bigr),\qquad
  u_c(\varepsilon)=-\frac{3\,h\,\varepsilon}{2D\,(\eta_c(\varepsilon)-\eta_c)}\bigl(1+O(|\varepsilon|^{1/3})\bigr).
\end{equation}
In particular $|\eta_c(\varepsilon)-\eta_c|\propto|h\varepsilon|^{2/3}$ and $|u_c(\varepsilon)|\propto|h\varepsilon|^{1/3}$. In the supercritical case $C<0$, $D<0$, the fold lies at $\eta_c(\varepsilon)<\eta_c$, i.e.\ the speciation is delayed along generation, and the mode connected continuously to the pre-speciation mode lies on the side $\operatorname{sgn}u=\operatorname{sgn}(\varepsilon h)$.
\end{proposition}
The exponent $2/3$ in \cref{eq:cusp} is the cusp law $|\Delta\eta_c|\propto|h|^{2/3}$ of an imperfect pitchfork: near a symmetric configuration the speciation level is much more sensitive to small errors than at a generic fold, since $|h|^{2/3}\gg|h|$ for small $|h|$. The law is observed in the reweighting experiment of~\Cref{sec:numerical}, with measured exponents $0.666$ and $0.662$. Moreover, the sign of $h$ alone decides which of the two would-be symmetric modes is connected continuously to the pre-speciation mode. For a reference field that is itself imperfect, with read-out $h_0\neq0$, this choice changes only if the error reverses the sign of the total read-out, i.e.\ if $|\varepsilon h|>|h_0|$, however large the transverse components of $\delta S$.
 
\medskip

\noindent \paragraph{4) Equilibria after the last speciation}
After the last speciation along a branch, the branch is a single mode, and the contraction becomes uniform in every direction. We state this result for an isotropic Gaussian mixture 
\begin{equation}
  p_0=\sum_{k=1}^K\alpha_k\,\mathcal N(\mu_k,s^2I_d),\qquad \Delta:=\max_{j,l}\|\mu_j-\mu_l\|,
\end{equation}
and we denote by $w_k(x,\eta)$ the posterior weights of the components.
 
\begin{corollary}[Contraction around a single mode]\label{cor:contraction}
Let $x^\star$ be a critical point at which one component has posterior weight $w_k(x^\star,\eta)\ge1-\epsilon$, and assume $\rho:=2\epsilon\Delta^2/(s^2+\eta)<1$. Then
\begin{equation}\label{eq:contraction}
  \|\delta x^\star\|\le\frac{s^2+\eta}{1-\rho}\,\|\delta S(x^\star)\|+O(\|\delta S\|^2).
\end{equation}
\end{corollary}
 
\begin{proof}
By \cref{prop:mixture-critical-bound}, $C_\eta=W_\eta+B_\eta$, with $W_\eta=\frac{\eta s^2}{s^2+\eta}I_d$. The component posterior means $m_j=(s^2x+\eta\mu_j)/(s^2+\eta)$ differ by at most $\eta\Delta/(s^2+\eta)$, and $m_\eta-m_k=\sum_lw_l(m_l-m_k)$ has norm at most $\epsilon\,\eta\Delta/(s^2+\eta)$. Hence
\[
  \|B_\eta\|\le\sum_jw_j\|m_j-m_\eta\|^2\le\bigl((1-\epsilon)\epsilon^2+\epsilon\bigr)\frac{\eta^2\Delta^2}{(s^2+\eta)^2}\le\frac{2\epsilon\,\eta^2\Delta^2}{(s^2+\eta)^2}.
\]
Therefore $H_\eta=(C_\eta-\eta I_d)/\eta^2=-(s^2+\eta)^{-1}\bigl(I_d-(s^2+\eta)B_\eta/\eta^2\bigr)$, with $\|(s^2+\eta)B_\eta/\eta^2\|\le\rho$, so that $\|H_\eta^{-1}\|\le(s^2+\eta)/(1-\rho)$. Conclude with $\delta x^\star=-H_\eta^{-1}\delta S(x^\star)$.
\end{proof}
 
Thus, beyond the last speciation, the committed configuration responds to the score error isotropically, with gain at most about $s^2+\eta$. This gain vanishes as $\eta\to0$ only for point masses ($s=0$), as for the empirical distribution; for components of width $s>0$ it saturates at the data scale $s^2$.
 
\begin{remark}\label{rem:contraction}
The mechanism is not specific to Gaussian mixtures: once the posterior is concentrated on a single log-concave cluster, $C_\eta(x^\star)\prec\eta I_d$ uniformly, $H_\eta(x^\star)$ is uniformly negative definite, and~\cref{prop:robust} applies with $\gamma_0$ bounded away from zero.
\end{remark}

\section{Numerical Experiments on the Fashion-MNIST dataset}\label{sec:numerical}

We test the local theory on pairs of Fashion-MNIST classes. We first use the exact score of the corresponding noised empirical distribution, and then we repeat the tests with scores learned by different neural networks.

\smallskip

\noindent The codes that implement the numerical experiments on the Fashion-MNIST dataset are provided in the \texttt{FashionMNIST\_Tests} folder of the repository accompanying this paper.

\medskip

\noindent \textit{Dataset.}
We consider two pairs of classes extracted from the Fashion-MNIST collection: Trouser/Sneaker, whose classes are well separated, and Pullover/Coat, whose classes overlap strongly.
For each pair $(c_1,c_2)$ we draw $n=1000$ training images per class, rescale the pixel values to $[0,1]$, subtract the mean of the pooled $2n$ images, and project onto the first $d=30$ principal directions of the pool. This yields points $y_1,\dots,y_N\in\mathbb{R}^d$, $N=2n$, with labels in $\{0,1\}$, and the data distribution
\begin{equation}\label{eq:fm-p0}
  p_0=\sum_{i=1}^N\omega_i\,\delta_{y_i},\qquad \omega_i=\frac1N .
\end{equation}
We apply the diffusion directly to the projected data in $\mathbb{R}^d$. While this is equivalent to adding Gaussian noise to the original images and then projecting them, the speciations we compute are those of the projected data and need not coincide with those of the full images.
However, we expect that the first 30 principal directions capture the most part of the data variation, including speciations. Namely, they account for $88\%$ of the total variance for Trouser/Sneaker and $85\%$ for Pullover/Coat, and they contain almost entirely the vector $\mu_{c_2}-\mu_{c_1}$ joining the two class means, i.e.\ the average image of class $c_2$ minus that of class $c_1$ ($100\%$ and $99.7\%$ of its squared norm, respectively), so that the direction separating the two classes on average lies in the projected space.

\smallskip

\noindent \textit{Locating candidate speciation points.}
By Remark~\ref{rem:empirical}, the posterior at $(x,\eta)$ is the discrete law with softmax weights $w_i(x,\eta)\propto\omega_i\exp(-\|x-y_i\|^2/(2\eta))$. The score $S_\eta(x)=(m_\eta(x)-x)/\eta$, the posterior covariance $C_\eta(x)$ and the cumulants entering~\cref{prop:coeff} are weighted sums over the $N$ points, as in \cref{rem:empirical}. 
Along the reverse process, as $\eta$ decreases, we look for the first change in the number of modes of $p_\eta$.
Modes are computed by the mean-shift iteration~\cite{fukunaga1975estimation,cheng1995mean} $x_{k+1}=m_\eta(x_k)$, which repeatedly applies the posterior-mean denoiser at fixed noise level. By Tweedie's formula, $m_\eta(x)=x+\eta\,S_\eta(x)$, so each step is a gradient-ascent step on $\log p_\eta$ with step size $\eta$; its fixed points are the critical points of $p_\eta$ (\cref{cor:critical}(i)). 
For the Gaussian kernel the iteration increases the density at every step: by Jensen's inequality,
\begin{equation}\label{eq:meanshift-monotone}
  \log p_\eta(x_{k+1})-\log p_\eta(x_k)\ \ge\ \frac{\|x_{k+1}-x_k\|^2}{2\eta},
\end{equation}
so that the steps vanish and every limit point is a critical point of $p_\eta$. Since saddles are unstable for this ascent, the limit is a mode for all but exceptional starting points; in the computations this is checked a posteriori on the spectrum of $H_\eta$.
Each run of the iteration converges to the mode whose basin of attraction contains the starting point. To find the modes at a given $\eta$, we therefore run it from $19$ starting points spread over the data: the two class means, the mean of the pool and $16$ randomly chosen training images. The distinct limits are the modes, and their number is recorded as a function of $\eta$. This procedure is not guaranteed to find every mode, but it is sufficient to detect the first value of $\eta$ at which a second mode appears near the classes; the speciation point is then computed independently of the starting points by solving \eqref{eq:fm-extended}.
The iteration is started from the two class means, the pool mean and $16$ random data points, and is refined by a few iterations of Newton's method on $S_\eta(x)=0$, whose Jacobian $H_\eta=(C_\eta-\eta I_d)/\eta^2$ is available in closed form. 
We scan $60$ geometrically spaced values of $\eta$ from $2\lambda_{\max}$ to $0.02\lambda_{\max}$, where $\lambda_{\max}$ is the largest eigenvalue of the covariance of the pool; at the upper end $p_\eta$ is unimodal. 
Let $\eta_-$ be the first value with two modes and $\eta_+$ the previous one. The saddle between the two modes is located by minimizing $\log p_\eta$ on the segment joining them and polishing with Newton's method; it is accepted only if exactly one eigenvalue of $C_\eta$ exceeds $\eta$, i.e.\ if it has exactly one unstable direction. The saddle is then continued towards $\eta_+$ by bisection in $\eta$, with Newton's method warm-started at each step, until it ceases to be a saddle. Finally, the soft critical point $(x_c,\eta_c)$ is refined by Newton's method on the extended system
\begin{equation}\label{eq:fm-extended}
  S_\eta(x)=0,\qquad \frac{\xi_{\max}(C_\eta(x))}{\eta}-1=0,
\end{equation}
which consists of $d+1$ equations in the $d+1$ unknowns $(x,\eta)$ and characterizes soft critical points by~\cref{cor:critical}(iii). The residual of~\cref{eq:fm-extended} at the solution is below $10^{-12}$.

\subsection{Exact score}
In this case we can compute exactly the quantities
involved in the estimates.
Namely, we compute at $(x_c,\eta_c)$:
\begin{enumerate}
\item[(i)] the \emph{spectral gap} $(\eta_c-\xi_2)/\eta_c$, where $\xi_2$ is the second eigenvalue of $C_{\eta_c}(x_c)$. A gap bounded away from zero confirms~\cref{ass:simple} and justifies the scalar reduction;
\item[(ii)] the hypothesis of~\cref{prop:mixture-critical-bound} with the two classes as components ($K=2$): for each class $j$ we compute the within-class posterior covariance $C_j$, obtained by renormalizing the weights $w_i$ over the points of class $j$, and the ratio $\lambda_{\max}(C_j)/\eta_c$. If both ratios are smaller than $1$, \cref{prop:mixture-critical-bound} gives $\dim\ker H_{\eta_c}(x_c)\le1$ rigorously. For non-Gaussian components this condition is not automatic and must be checked;
\item[(iii)] the alignment $|\langle e_c,\hat\mu\rangle|$ of the critical direction with the unit vector $\hat\mu$ joining the two class means, and the posterior mass of class $c_2$ at the two modes found at $\eta_-$. These quantities tell whether the speciation separates the classes;
\item[(iv)] the reduced coefficients $A,B,C,D$ of \eqref{eq:ABC}, from the closed formulas of ~\cref{prop:coeff}. The critical direction is oriented so that $\langle e_c,\hat\mu\rangle>0$; this fixes the signs of $A$ and $B$, while $AB$, $C$ and $D$ do not depend on the orientation. Uncertainty is quantified by a nonparametric bootstrap: the $n$ images of each class are resampled with replacement ($100$ resamples), the soft critical point is recomputed by solving \eqref{eq:fm-extended} from $(x_c,\eta_c)$, and the coefficients are recomputed there. We report $95\%$ percentile intervals. By ~\cref{prop:fold}, if $A$ and $B$ are significantly different from zero the speciation is a nondegenerate fold, and the new mode--saddle pair is created along the reverse process if and only if $AB>0$.
\end{enumerate}

\medskip

\noindent As a control with an exact $\mathbb Z_2$ symmetry, we add to the pool the reflected points $2c-y_i$, where $c$ is the midpoint of the two class means. The resulting measure is invariant under the point reflection $R:x\mapsto2c-x$, so (P0) holds with $x_c=c$ for every critical direction. By ~\cref{lem:sym}, $c$ is a critical point for every $\eta$ and $A=B=0$. The pitchfork occurs at the value $\eta_c^{\rm sym}$ solving $\xi_{\max}(C_\eta(c))=\eta$, computed by bisection, and~\cref{prop:pitchfork} predicts a supercritical speciation if $C<0$ and $D<0$. For the bootstrap of the control ($20$ resamples), each resample of the original pool is symmetrized in the same way.

\medskip

\noindent \textit{Metrics.}
We test three predictions of the reduced equations:
\begin{enumerate}
\item[(i)] \emph{Barrier exponents.} For $|\eta-\eta_c|=s\,\eta_c$ with $14$ values $s\in[10^{-4},5\cdot10^{-2}]$, on the side of $\eta_c$ where the new critical points exist, we compute the new mode and the saddle by Newton's method, starting from the leading-order positions $x_c\pm u_\pm e_c$ given by \eqref{eq:foldbranches} or \eqref{eq:pitchbranches}, check their types, and measure the barrier $\Delta\varphi=|\log p_\eta(x_{\rm mode})-\log p_\eta(x_{\rm saddle})|$. Remarks~\ref{rem:foldbarrier} and~\ref{rem:pitchbarrier} predict $\Delta\varphi\simeq\frac{4\sqrt2}{3}|B|^{3/2}|A|^{-1/2}|\eta-\eta_c|^{3/2}$ for the fold and $\Delta\varphi\simeq\frac32\frac{D^2}{|C|}(\eta-\eta_c)^2$ for the pitchfork. These include the prefactors, which are computed independently from~\cref{prop:coeff}. The exponent is estimated by a least-squares fit in log-log scale on the six smallest values of $s$.
\item[(ii)] \emph{Cusp law.} Starting from the symmetrized pool, we break the symmetry in a controlled way by assigning weight $1+2\alpha$ to the original points and $1-2\alpha$ to the reflected ones. With $h(\alpha):=e_c^\top S_{\eta_c^{\rm sym}}(c)$, the imperfect pitchfork $\Phi_\eta(u)=h+D\chi u+\frac C6u^3$ has a single fold, where $\Phi=\partial_u\Phi=0$. Eliminating $u$ gives
\begin{equation}\label{eq:fm-cusp}
  \chi^3=-\frac{9\,C\,h^2}{8\,D^3},\qquad u=-\frac{3h}{2D\chi},
\end{equation}
so that $|\eta_c(\alpha)-\eta_c^{\rm sym}|\propto|h|^{2/3}\propto\alpha^{2/3}$. For $9$ values $\alpha\in[10^{-4},3\cdot10^{-2}]$ we solve \eqref{eq:fm-extended} for the reweighted measure, using \eqref{eq:fm-cusp} only as initial guess, and fit the exponent on the five smallest values of $\alpha$.
\item[(iii)] \emph{Critical read-out.} We perturb the exact score by a constant field, $S_\eta\mapsto S_\eta+\varepsilon v$, and locate the perturbed fold by solving \eqref{eq:fm-extended} with $S_\eta$ replaced by $S_\eta+\varepsilon v$. We take $v=e_c$ and $v=v_\perp$, a random unit vector of $E_s$, and $7$ values $|\varepsilon|\le0.02\,\eta_c|B|$. \Cref{prop:foldpersist} predicts $\eta_c(\varepsilon)-\eta_c=-\varepsilon\langle e_c,v\rangle/B+O(\varepsilon^2)$: slope $-1/B$ for $v=e_c$ and no first-order shift for $v=v_\perp$.
\end{enumerate}

\begin{table}[ht]
\centering\small
\renewcommand{\arraystretch}{1.15}
\begin{tabular}{lcc}
\hline
 & Trouser/Sneaker & Pullover/Coat\\
\hline
$\eta_c$ ($\sigma_c=\sqrt{\eta_c}$) & $20.54$ ($4.53$) & $6.76$ ($2.60$)\\
spectral gap $(\eta_c-\xi_2)/\eta_c$ & $0.85$ & $0.62$\\
$\lambda_{\max}(C_j)/\eta_c$, classes $c_1$, $c_2$ & $0.22$, $0.22$ & $0.93$, $1.08$\\
$|\langle e_c,\hat\mu\rangle|$ & $0.996$ & $0.23$\\
mass of class $c_2$ at the new modes & $0.26$, $0.82$ & $0.60$, $0.40$\\
$A$ & $7.4\cdot10^{-3}$ $[6.5,8.1]\cdot10^{-3}$ & $1.9\cdot10^{-2}$ $[1.5,2.4]\cdot10^{-2}$\\
$B$ & $3.8\cdot10^{-3}$ $[3.4,4.1]\cdot10^{-3}$ & $2.2\cdot10^{-2}$ $[1.9,2.6]\cdot10^{-2}$\\
bifurcation type & fold, $AB>0$ & fold, $AB>0$\\
\hline
symmetrized: $\eta_c^{\rm sym}$ & $24.05$ & $6.77$\\
symmetrized: $|A|,|B|$ & $<10^{-17}$ & $<10^{-16}$\\
symmetrized: $C$ & $-3.13\cdot10^{-3}$ $[-3.20,-3.01]\cdot10^{-3}$ & $-1.14\cdot10^{-2}$ $[-1.32,-0.95]\cdot10^{-2}$\\
symmetrized: $D$ & $-1.59\cdot10^{-3}$ $[-1.63,-1.53]\cdot10^{-3}$ & $-7.8\cdot10^{-3}$ $[-9.1,-6.7]\cdot10^{-3}$\\
\hline
barrier exponent, fold / pitchfork & $1.501$ / $2.000$ & $1.500$ / $2.000$\\
cusp exponent ($2/3$ expected) & $0.666$ & $0.662$\\
read-out slope, measured / $-1/B$ & $-264.8$ / $-263.9$ & $-46.04$ / $-46.03$\\
max transverse / critical shift & $2\cdot10^{-5}$ & $5\cdot10^{-5}$\\
\hline
\end{tabular}
\caption{Speciation diagnostics for two Fashion-MNIST class pairs ($n=1000$ images per class, PCA dimension $d=30$, exact empirical score). Brackets: $95\%$ bootstrap intervals. The quantities in the lower block are defined in items (a)--(c).}
\label{tab:fmnist}
\end{table}

\begin{figure}[t]
\centering
\includegraphics[width=1.\textwidth]{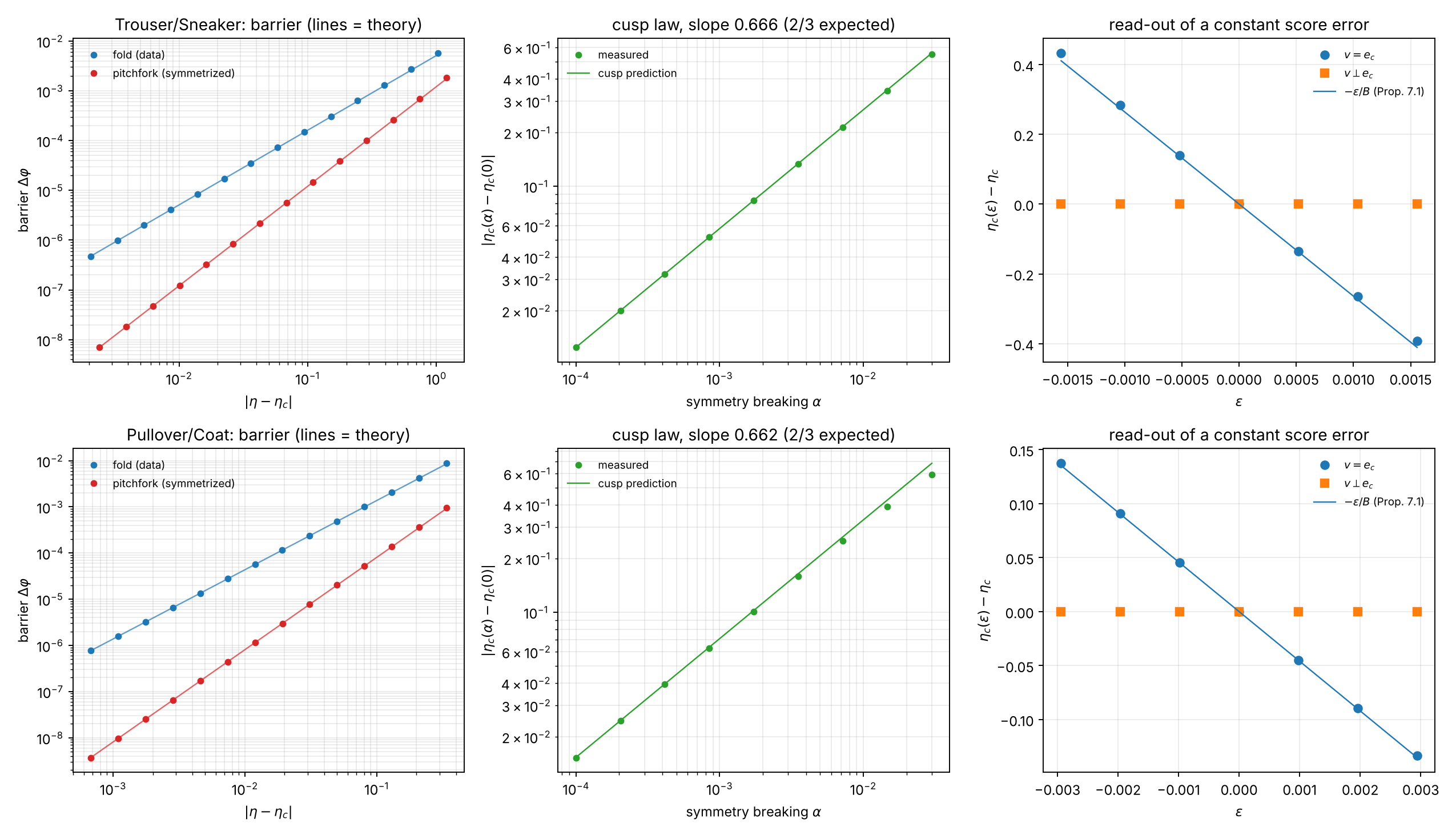}
\caption{Fashion-MNIST test case; top row Trouser/Sneaker, bottom row Pullover/Coat. Left: log-density barrier between the new mode and the saddle versus $|\eta-\eta_c|$, for the data (fold) and for the symmetrized pool (pitchfork); dots are computed critical points, lines are the predictions of Remarks~\ref{rem:foldbarrier} and~\ref{rem:pitchbarrier} with coefficients from~\cref{prop:coeff}. Center: shift of the speciation level under the symmetry-breaking reweighting $1\pm2\alpha$, compared with the cusp prediction~\cref{eq:fm-cusp}. Right: shift of the fold under a constant score error $\varepsilon v$, for $v=e_c$ and $v\perp e_c$, compared with~\cref{prop:foldpersist}.}
\label{fig:fmnist}
\end{figure}

\medskip

\noindent The results are reported in~\cref{tab:fmnist} and \cref{fig:fmnist}, and can be summarized as follows: 
\begin{itemize}
\item[a)] the speciation of class pairs from a realistic dataset is a fold: for both pairs, $A$ and $B$ are positive and their bootstrap intervals are well separated from zero. By~\cref{prop:fold} the first speciation is therefore a nondegenerate fold, and since $AB>0$ the new mode--saddle pair is created as the noise decreases, i.e.\ along generation. The asymmetry of the two classes, in their shape and in their position relative to the rest of the pool, is enough to unfold the pitchfork. In the symmetrized control the odd coefficients vanish to machine precision, as predicted by~\cref{lem:sym}, and $C<0$, $D<0$ with intervals excluding zero: the speciation is a supercritical pitchfork, in which a mode splits into two. The pitchfork is thus recovered only when the symmetry is imposed exactly.

\item[b)]the critical subspace is one-dimensional. In both cases the spectral gap is large, $0.85$ and $0.62$: the second eigenvalue of $C_{\eta_c}(x_c)$ is far below the threshold, and \cref{ass:simple} holds with a comfortable margin. For Trouser/Sneaker the hypothesis of \cref{prop:mixture-critical-bound} holds as well, since both within-class ratios are about $0.22$. The simplicity of the speciation is therefore not merely generic but guaranteed: no single class can create a critical direction on its own, and the only critical direction comes from the competition between the two classes. Consistently, $e_c$ is almost parallel to the class-mean direction, and the two new modes are dominated by different classes.

\item[c)] for overlapping classes, the first speciation is not a class split. In the Pullover/Coat pair, the critical direction is nearly orthogonal to the class-mean direction, $|\langle e_c,\hat\mu\rangle|=0.23$, both new modes contain a substantial fraction of both classes, and the within-class ratio of Coat exceeds one. The hypothesis of \cref{prop:mixture-critical-bound} therefore fails: the variability inside the Coat class is, at this noise level, larger than the separation between the classes, and the density first splits along a within-class direction. This is consistent with the theory, which determines when and along which direction the density splits, not whether the split is aligned with the labels. It also shows that the diagnostics (ii) and (iii) can tell a priori whether a speciation separates the classes

\item[d)]the barrier exponents are $1.50$ for the folds and $2.00$ for the pitchforks, and the measured barriers coincide with the predicted ones, prefactors included, over more than two decades of $|\eta-\eta_c|$ (\Cref{fig:fmnist}, left). The reweighting experiment reproduces the cusp law with exponents $0.666$ and $0.662$; for Pullover/Coat the largest values of $\alpha$ fall below the leading-order prediction, as expected outside the small-imperfection regime (\Cref{fig:fmnist}, center). Finally, the shift of the speciation level under a constant score error follows the critical read-out of ~\cref{prop:foldpersist} to within $0.3\%$, while an error of the same size along a transverse direction produces a shift about $10^{-5}$ times smaller, i.e.\ no shift at first order (\Cref{fig:fmnist}, right).
\end{itemize}

\subsection{Learned score}
We train six small networks (described in detail in Appendix) to learn the score $S_\theta$
via denoising score matching on the 
on the Trouser/Sneaker pair: the difference
\begin{equation}\label{eq:fm-dS}
  \delta S_\eta(x):=S_\theta(x,\eta)-S_\eta(x)
\end{equation}
is thus the approximation error of the network.
% and compare their speciation with the exact one ($\eta_c=20.54$, Table~\ref{tab:fmnist}). 
The Jacobian $J_\theta:=D_xS_\theta$ is computed exactly by backpropagation. It is not symmetric, so the learned field has distinct right and left critical vectors $v$ and $v^\ast$. The speciation of the learned field is therefore defined as a fold of its zeros: a point where $S_\theta(x,\eta)=0$ and the eigenvalue of $J_\theta$ with the largest real part vanishes. The procedure for locating the fold is the same as in the exact case.

\smallskip

\noindent \emph{Predictions.} At first order, \cref{prop:foldpersist} predicts the shift
\begin{equation}\label{eq:fm-first}
  \eta_c^\theta-\eta_c\simeq-\frac{h}{B},\qquad h:=e_c^\top\delta S_{\eta_c}(x_c),
\end{equation}
which depends on the error only through its critical read-out $h$, evaluated at the exact speciation point. A refined prediction keeps the reduced equation~\cref{eq:Phi-joint} beyond first order and adds the error together with its slope along $e_c$, $h':=e_c^\top D_x\delta S_{\eta_c}(x_c)\,e_c$:
\begin{equation}\label{eq:fm-reduced}
  h+h'u+B\chi+D\chi u+\frac A2u^2+\frac C6u^3=0,\qquad h'+D\chi+Au+\frac C2u^2=0 .
\end{equation}
Its solution $(u,\chi)$ gives the fold of the perturbed reduced problem, and hence the predicted shift $\chi$. The transverse part $\|P_s\delta S_{\eta_c}(x_c)\|$ of the error enters neither prediction.

\begin{table}[t]
\centering\small
\renewcommand{\arraystretch}{1.15}
\begin{tabular}{lcccccc}
\hline
network & rel.\ error & $|h|$ & $\|P_s\delta S\|$ & measured & first order \eqref{eq:fm-first} & reduced \eqref{eq:fm-reduced}\\
\hline
 N($64$, $2\,000$) & $4.4\%$ & $0.044$ & $0.027$ & no speciation & $-11.5$ & $-10.0$\\
N($64$, $8\,000$)  & $5.0\%$ & $0.058$ & $0.009$ & no speciation & $-15.4$ & $-15.0$\\
N($128$, $6\,000$)  & $1.1\%$ & $0.0012$ & $0.013$ & $+0.30$ & $-0.31$ & $+0.33$\\
N($128$, $15\,000$) & $0.7\%$ & $0.0036$ & $0.008$ & $+1.26$ & $+0.94$ & $+2.19$\\
N($256$, $8\,000$) & $1.1\%$ & $0.0070$ & $0.012$ & $+2.37$ & $+1.86$ & $+2.19$\\
N($256$, $16\,000$) & $0.7\%$ & $0.0022$ & $0.008$ & $-0.49$ & $-0.57$ & $-0.57$\\
\hline
\end{tabular}
\caption{Speciation of learned scores for Trouser/Sneaker (exact value $\eta_c=20.54$, $B=3.79\cdot10^{-3}$) obtained by 6 different networks N(width, \#steps), with 2000 steps corresponding to about $5\cdot 10^2$ passes over the images.
The relative error is $\|\delta S_{\eta_c}(x_c)\|$ divided by the average norm of the exact score at noise level $\eta_c$; $h=e_c^\top\delta S_{\eta_c}(x_c)$ is the critical read-out and $\|P_s\delta S\|$ the transverse part of the error at the exact speciation point. The last three columns give the shift $\eta_c^\theta-\eta_c$. ``No speciation'': the learned density has a single mode for all $\eta\in[\eta_c/3,3\eta_c]$.}
\label{tab:fmnist-learned}
\end{table}

\begin{figure}[t]
\centering
\includegraphics[width=\textwidth]{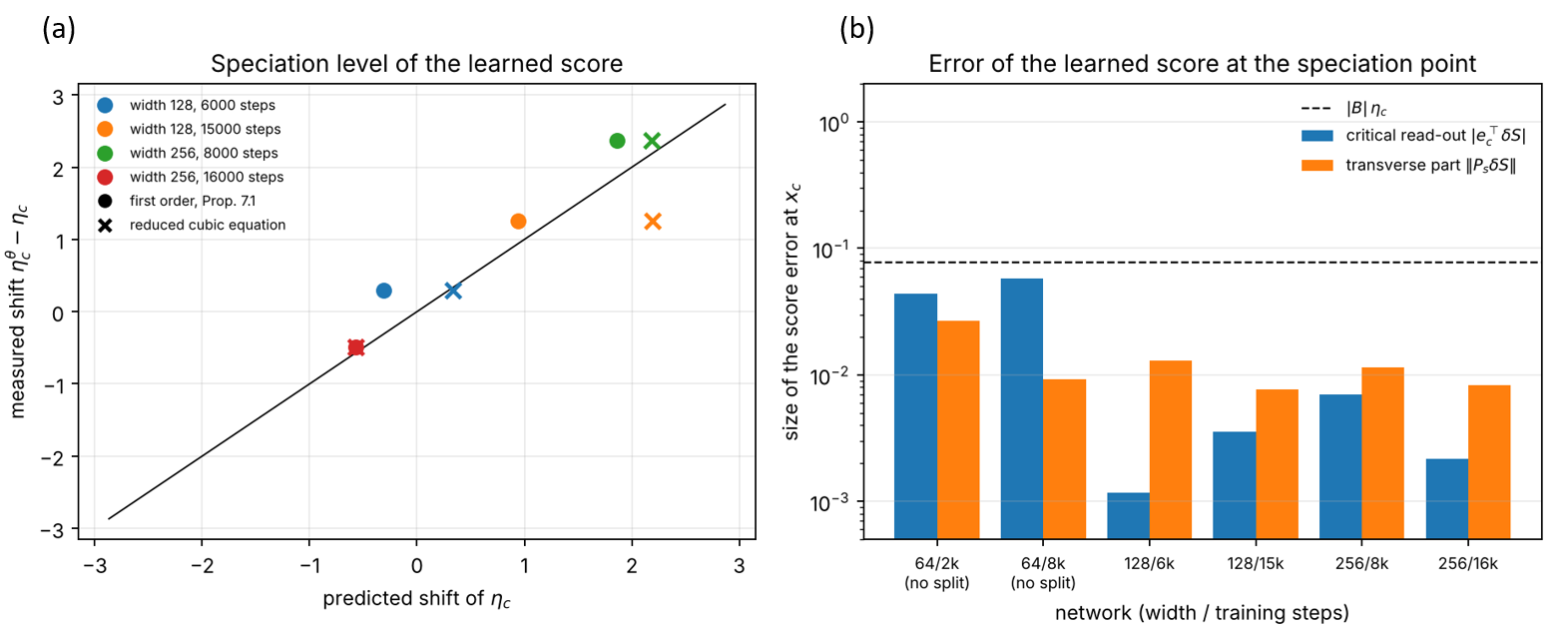}
\caption{Learned scores, Trouser/Sneaker. (a): measured shift of the speciation level versus the first-order prediction \eqref{eq:fm-first} (dots) and the prediction of the reduced equation \eqref{eq:fm-reduced} (crosses), for the four networks that speciate; the line is the identity; (b): critical read-out $|h|$ and transverse part $\|P_s\delta S\|$ of the error at the exact speciation point, for all six networks; the dashed line is $|B|\eta_c$, the read-out that would shift $\eta_c$ by $\eta_c$ itself at first order.}
\label{fig:fmnist-learned}
\end{figure}

\smallskip

\noindent {\textit{Results}}. The results are reported in~\Cref{tab:fmnist-learned} and ~\Cref{fig:fmnist-learned}, and can be summarized as follows:
\begin{itemize}
    
\item[(e)] only the critical component of the error matters. The shift of the speciation level follows the read-out $h$, and is insensitive to the transverse part of the error, which is comparable to or larger than $|h|$ for all the networks that speciate, and up to ten times larger for the network of width $128$ trained for $6\,000$ steps
\item[(f)] the factor $1/B\approx264$ plays the role of a condition number which impacts on the shift of $\eta_c$. The two smallest networks have a relative error of only $4$--$5\%$ on the score, but their error is concentrated along $e_c$, with a read-out of $0.56$--$0.75$ times $|B|\eta_c$. Their learned density remains unimodal over the whole training range: they do not separate the two classes at all, smoothing out the bimodal structure
% \item[(g)] the reduced description is quantitatively predictive. For the two networks of width $256$, the first-order prediction differs from the measured shift by about $15$--$25\%$, and the reduced equation \eqref{eq:fm-reduced} by about $8$--$16\%$. For the network of width $128$ trained for $6\,000$ steps, the read-out is so small that the first-order prediction even has the wrong sign; the reduced equation, which includes the slope $h'$ of the error along $e_c$, recovers the measured shift. For the network of width $128$ trained for $15\,000$ steps the first-order and the reduced predictions deviate by about $25\%$ and $75\%$, respectively, which marks the limit of the local regime. For the two smallest networks the predicted shifts are a large fraction of $\eta_c$, outside the domain of validity of the local reduction, and we do not attempt a quantitative comparison.

\item[(g)] the learned fields are not gradients, but only mildly so: at the learned speciation the relative asymmetry $\|J_\theta-J_\theta^\top\|_F/\|J_\theta\|_F$ is between $1\%$ and $2.4\%$, the angle between the right and left critical vectors $v$ and $v^\ast$ is between $1.7^\circ$ and $5.1^\circ$, and $v$ is aligned with the exact critical direction ($|\langle v,e_c\rangle|/\|v\|>0.999$)
\end{itemize}

\section{Discussion and conclusions}
\label{sec:discussion}

In this work we have developed a local theory of speciation in generative diffusion models, based on the bifurcations of the critical set of the noised density. 
At a speciation, the Hessian of the noised log-density loses rank, which happens exactly when the posterior variance of the clean datum along some direction $e_c$ reaches the accumulated variance $\eta$. For the case of a unique critical direction, we used the Lyapunov--Schmidt reduction to eliminate the uniformly stable directions and leave a scalar equation for the critical amplitude $u$,
\begin{equation}\label{eq:concl-reduced}
  0=B\,(\eta-\eta_c)+D\,(\eta-\eta_c)\,u+\frac A2\,u^2+\frac C6\,u^3+\cdots,
\end{equation}
whose coefficients are explicit posterior cumulants and can be computed exactly for empirical data.  In the generic case, $A,B\neq0$, and the speciation is a nondegenerate fold: a new mode is born together with a saddle, rather than an existing mode splitting into two. The pitchfork, in which a mode splits into two symmetric modes, requires $A=B=0$, which is enforced by an exact reflection symmetry of the data, and is unfolded by any symmetry-breaking perturbation. Simple speciations are guaranteed for two-component Gaussian mixtures; higher-dimensional critical subspaces require additional structure.

\medskip

\noindent The reduction also identifies the role of errors in the representation of the score. To first order, a score error $\delta S$ affects a simple speciation only through its projection
onto the critical direction. At a generic fold the speciation level shifts linearly with the critical component of the score mediated by a conditioning gain factor $-1/B$, which can cause a significant increase in the shift. The transverse components of the error are absorbed by the spectral gap: they reach the critical amplitude only through the tilt of the slaving manifold, which vanishes at the speciation point. For a learned, non-gradient score the critical direction is replaced by the left critical eigenvector. The experiments on Fashion-MNIST show that, with learned scores, the speciation level follows the critical component of the network error rather than its global size, and small networks with a few percent of global error fail to separate the classes, because their error is concentrated precisely along the critical direction. The experiments in this paper are completed and integrated 
by those carried out in the companion paper~\cite{softmodes2026}.

\medskip

\noindent The present analysis has limits, which also open directions for further research work. Namely:
\begin{enumerate}[label=(\roman*)]
\item Critical subspaces of dimension $>1$. When several classes separate simultaneously, the posterior covariance has a multiple eigenvalue equal to $\eta_c$. The Lyapunov--Schmidt reduction then yields a multidimensional bifurcation equation, whose analysis falls within equivariant bifurcation theory~\cite{GolubitskySchaeffer1985}
\item Frozen noise level versus reverse dynamics. Our analysis concerns the score field at frozen noise level. Along the reverse process, noise levels vary continuously, and the trajectory commits to a branch dependently on the sampler and on the schedule. Relating the local events described here to the global sensitivity of the generated sample to perturbations/guidance \cite{georgiev2023journey,kynkaanniemi2024applying} along the whole trajectory requires an analysis of the non-autonomous dynamics.
\item {Scale of the experiments.} The experiments concern pairs of classes in a principal subspace and small networks trained around the speciation level. Measuring the critical read-out of large networks trained over the full range of noise levels, and on full-dimensional data, is a natural interesting next step.
\end{enumerate}

\section*{Acknowledgments}
This research has been partially performed in the framework of the GNCS Project 
\textit{``Oltre il gradiente deterministico: a\-na\-li\-si di dinamiche stocastiche per funzioni non convesse"}, CUP: E53C25002010001. 
Part of the computational resources were provided by the INDACO Platform, a project of High Performance Computing of the University of Milan ({\tt https://www.indaco.unimi.it/}).

\bibliographystyle{siamplain}
\bibliography{biblio}

%%%%%%%%%%%%%%%%%%%%%%%
\appendix

\section{Technical proofs}
\crefalias{section}{appendix}
\crefalias{subsection}{appendix}
\label{sec:proofs}

This sections collects technical proofs that are not reported in the main text for readability. 

\subsection{Proof of~\cref{lem:reduced}}
\label{app:lem:reduced}
Fix $\eta\in I_c$, write $x=x(u,\eta)$, and let
  $w:=\partial_u x(u,\eta)=e_c+V_\eta'(u)$.
Since $V_\eta$ takes values in $E_s=e_c^\perp$, also $V_\eta'(u)\in E_s$. Hence
\begin{equation}\label{eq:w-props}
  e_c^\top w=1,\qquad S_\eta(x)^\top V_\eta'(u)=\bigl(P_sS_\eta(x)\bigr)^\top V_\eta'(u).
\end{equation}

(i) By the chain rule and \eqref{eq:w-props},
\begin{equation}    
  \widetilde U_\eta'(u)=S_\eta(x)^\top w=e_c^\top S_\eta(x)+\bigl(P_sS_\eta(x)\bigr)^\top V_\eta'(u).
\end{equation}
The second term vanishes, because $V_\eta$ is defined by $P_sS_\eta(x(u,\eta))\equiv0$, and the first is $\Phi(u,\eta)$ by \eqref{eq:bif}. Thus $\widetilde U_\eta'=\Phi(\cdot,\eta)$, and the claim on the critical points follows from~\cref{thm:LS}.

(ii) Differentiating the identity $P_sS_\eta(x(u,\eta))\equiv0$ in $u$ gives $P_sH_\eta(x)\,w=0$, so $H_\eta(x)\,w$ is parallel to $e_c$. Differentiating $\widetilde U_\eta'(u)=e_c^\top S_\eta(x(u,\eta))$ in $u$ gives $e_c^\top H_\eta(x)\,w=\widetilde U_\eta''(u)$. Therefore
\begin{equation}\label{eq:Hw}
  H_\eta(x)\,w=\widetilde U_\eta''(u)\,e_c .
\end{equation}
Since $H_\eta(x)$ is symmetric, \eqref{eq:Hw} and \eqref{eq:w-props} give, for every $z\in E_s$,
\[
  w^\top H_\eta(x)\,z=\bigl(H_\eta(x)\,w\bigr)^\top z=\widetilde U_\eta''(u)\,e_c^\top z=0,
  \qquad
  w^\top H_\eta(x)\,w=\widetilde U_\eta''(u)\,e_c^\top w=\widetilde U_\eta''(u).
\]
Because $e_c^\top w=1$, the vector $w$ does not lie in $E_s$, so $\{w\}\cup E_s$ spans $\mathbb{R}^d$. In this basis the quadratic form of $H_\eta(x)$ is block diagonal, with blocks $\widetilde U_\eta''(u)$ and $N:=P_sH_\eta(x)|_{E_s}$. At $(u,\eta)=(0,\eta_c)$ we have $N=H_c|_{E_s}$, which is negative definite by~\cref{lem:split}(iii). By continuity, $N$ remains negative definite after possibly shrinking $\varepsilon$ and $\delta$. By Sylvester's law of inertia, $H_\eta(x)$ has the same number of positive, zero and negative eigenvalues as $\operatorname{diag}(\widetilde U_\eta''(u),N)$. Hence at a critical point $x(u_*,\eta)$, all eigenvalues of $H_\eta$ are negative if $\widetilde U_\eta''(u_*)<0$, and exactly one is positive if $\widetilde U_\eta''(u_*)>0$.

\subsection{Proof of~\cref{prop:coeff}} 
\label{app:prop:coeff}
Throughout, $\varphi=\varphi_{\eta_c}$, $H=\nabla^2\varphi$, and all derivatives are taken at $(u,\eta)=(0,\eta_c)$, where $x=x_c$, $\nabla\varphi(x_c)=0$, $e_c^\top H_c=0$ and $\partial_ux=e_c$. By ~\cref{lem:cumulants} and \eqref{eq:Hsinv},
\begin{equation}\label{eq:dict-proof}
  \nabla^r\varphi(x_c)=\frac{\kappa_r}{\eta_c^r}\quad(r\ge3),\qquad
  \bigl(H_c|_{E_s}\bigr)^{-1}=\eta_c^2\,R_s .
\end{equation}

\emph{Coefficients $A$ and $C$.} Fix $\eta=\eta_c$ and write $x(u)=x(u,\eta_c)$, so that $x'=e_c+V_{\eta_c}'$, while $x''=V_{\eta_c}''$ and $x'''=V_{\eta_c}'''$ lie in $E_s$. Differentiating $\Phi_{\eta_c}(u)=e_c^\top\nabla\varphi(x(u))$,
\[
  \partial_u^2\Phi=e_c^\top\bigl[\nabla^3\varphi(x',x',\cdot\,)+H\,x''\bigr],\qquad
  \partial_u^3\Phi=e_c^\top\bigl[\nabla^4\varphi(x',x',x',\cdot\,)+3\,\nabla^3\varphi(x',x'',\cdot\,)+H\,x'''\bigr].
\]
At $u=0$ the terms with $H$ vanish, so $A=\nabla^3\varphi(x_c)(e_c,e_c,e_c)$, which is \eqref{eq:A}, and
\[
  C=\nabla^4\varphi(x_c)(e_c,e_c,e_c,e_c)+3\,\nabla^3\varphi(x_c)\bigl(e_c,e_c,V_{\eta_c}''(0)\bigr).
\]
Differentiating $P_sS_{\eta_c}(x(u))\equiv0$ twice gives $P_s\bigl[\nabla^3\varphi(x',x',\cdot\,)+H\,x''\bigr]=0$, which at $u=0$, since $P_sH_c=H_c$, reads
\[
  V_{\eta_c}''(0)=-\bigl(H_c|_{E_s}\bigr)^{-1}P_s\nabla^3\varphi(x_c)(e_c,e_c,\cdot\,)=-\frac{1}{\eta_c}\,R_s\,k_c .
\]
Substituting, and using \eqref{eq:dict-proof} once more, gives \eqref{eq:C}.

\medskip

\noindent \emph{Noise derivatives.} By ~\cref{lem:noise}(i), $\partial_\eta\nabla\varphi_\eta=\tfrac12\nabla\Delta_x\varphi_\eta+H_\eta\nabla\varphi_\eta$, and differentiating once more,
\[
  e_c^\top\partial_\eta H_\eta\,e_c=\tfrac12\,\partial_{e_c}^2\Delta_x\varphi_\eta+\|H_\eta e_c\|^2+\bigl\langle\nabla\varphi_\eta,\nabla^3\varphi_\eta(e_c,e_c,\cdot\,)\bigr\rangle .
\]
At $(x_c,\eta_c)$ all terms containing $\nabla\varphi$ or $H_ce_c$ vanish. Since $\Delta_x\varphi_\eta=\eta^{-2}\operatorname{tr}C_\eta-d/\eta$ by \eqref{eq:tweedie-hess}, two applications of \eqref{eq:dercum} give $\nabla\Delta_x\varphi=\eta_c^{-3}m_3$ and $\partial_{e_c}^2\Delta_x\varphi=\eta_c^{-4}\sum_j\kappa_4(e_c,e_c,e_j,e_j)$. Hence
\begin{equation}\label{eq:noise-derivs}
  \partial_\eta S_\eta(x_c)\big|_{\eta_c}=\frac{m_3}{2\eta_c^3},\qquad
  e_c^\top\partial_\eta H_\eta(x_c)\big|_{\eta_c}\,e_c=\frac{1}{2\eta_c^4}\sum_{j=1}^d\kappa_4(e_c,e_c,e_j,e_j).
\end{equation}
Let $\dot x:=\partial_\eta V_\eta(0)|_{\eta_c}\in E_s$. Differentiating $P_sS_\eta(x(0,\eta))\equiv0$ in $\eta$ gives $P_s\bigl[\partial_\eta S_\eta(x_c)+H_c\,\dot x\bigr]=0$, so that, by \eqref{eq:noise-derivs},
\[
  \dot x=-\bigl(H_c|_{E_s}\bigr)^{-1}P_s\,\partial_\eta S_\eta(x_c)=-\frac{1}{2\eta_c}\,R_s\,P_sm_3 .
\]

\emph{Coefficient $B$.} Since $\Phi_\eta(0)=e_c^\top S_\eta(x(0,\eta))$ and $e_c^\top H_c=0$, we get $B=e_c^\top\partial_\eta S_\eta(x_c)$, which is \eqref{eq:B} by \eqref{eq:noise-derivs}. The second expression in \eqref{eq:B} follows from $\langle v,m_3\rangle=M_v$.

\emph{Coefficient $D$.} We have $\partial_u\Phi_\eta(0)=e_c^\top H_\eta(x(0,\eta))\,w(\eta)$, with $w(\eta):=e_c+V_\eta'(0)$ and $w(\eta_c)=e_c$. Differentiating in $\eta$ at $\eta_c$, the term containing $\partial_\eta w$ vanishes because $e_c^\top H_c=0$, and
\[
  D=e_c^\top\partial_\eta H_\eta(x_c)\,e_c+\nabla^3\varphi(x_c)(e_c,e_c,\dot x).
\]
Substituting \eqref{eq:noise-derivs}, the expression of $\dot x$ and \eqref{eq:dict-proof} gives \eqref{eq:D}.

\subsection{Proof of~\cref{prop:foldpersist}}
\label{app:prop:foldpersist}
The proof of~\cref{thm:LS} does not use the gradient structure of the field, only the invertibility of $H_c|_{E_s}$. It therefore applies to $S_\eta+\varepsilon\,\delta S_\eta$, with $\varepsilon$ as an additional parameter and the same splitting $\mathbb{R}^d=\operatorname{span}\{e_c\}\oplus E_s$. This gives a slaving map $V_{\eta,\varepsilon}(u)$, with $V_{\eta,0}=V_\eta$, and a reduced function
\[
  \Phi_{\eta,\varepsilon}(u):=e_c^\top\bigl(S_\eta+\varepsilon\,\delta S_\eta\bigr)\bigl(x_c+u\,e_c+V_{\eta,\varepsilon}(u)\bigr),\qquad \Phi_{\eta,0}=\Phi_\eta .
\]
At $(u,\eta,\varepsilon)=(0,\eta_c,0)$ the dependence of $V_{\eta,\varepsilon}$ on $\varepsilon$ contributes $e_c^\top H_c\,\partial_\varepsilon V=0$, so that
\begin{equation}\label{eq:dPhi-eps}
  \partial_\varepsilon\Phi_{\eta,\varepsilon}(u)\big|_{(0,\eta_c,0)}=h .
\end{equation}
A fold of the perturbed field is a solution $(u,\eta)$ of $\Phi_{\eta,\varepsilon}(u)=0$, $\partial_u\Phi_{\eta,\varepsilon}(u)=0$. At $(0,\eta_c,0)$ the Jacobian of this system with respect to $(u,\eta)$ is, by \cref{eq:ABC},
\begin{equation}  \begin{pmatrix}\partial_u\Phi&\partial_\eta\Phi\\ \partial_u^2\Phi&\partial_\eta\partial_u\Phi\end{pmatrix}
  =\begin{pmatrix}0&B\\ A&D\end{pmatrix}
\end{equation}
with determinant $-AB\neq0$ by (F1)--(F2). The implicit function theorem gives a unique $C^1$ solution $(u_c(\varepsilon),\eta_c(\varepsilon))$ with $(u_c(0),\eta_c(0))=(0,\eta_c)$, and we set $x_c(\varepsilon):=x_c+u_c(\varepsilon)\,e_c+V_{\eta_c(\varepsilon),\varepsilon}(u_c(\varepsilon))$. The fold is nondegenerate because the perturbed coefficients $A$ and $B$ remain nonzero by continuity, and the transverse block remains invertible.
 
\emph{Shift in $\eta$.} Differentiating $\Phi_{\eta_c(\varepsilon),\varepsilon}(u_c(\varepsilon))=0$ at $\varepsilon=0$, and using $\partial_u\Phi=0$ there and \eqref{eq:dPhi-eps}, gives $B\,\eta_c'(0)+h=0$, i.e.\ \eqref{eq:foldshift}.
 
\emph{Shift in space.} Differentiating $\partial_u\Phi_{\eta_c(\varepsilon),\varepsilon}(u_c(\varepsilon))=0$ at $\varepsilon=0$ gives $A\,u_c'(0)+D\,\eta_c'(0)+\partial_\varepsilon\partial_u\Phi=0$, where $\partial_\varepsilon\partial_u\Phi=h'+\nabla^3\varphi_{\eta_c}(x_c)(e_c,e_c,\partial_\varepsilon V)$ at $(0,\eta_c,0)$. Differentiating $P_s(S_\eta+\varepsilon\,\delta S_\eta)(x_c+V_{\eta,\varepsilon}(0))\equiv0$ in $\varepsilon$ gives $\partial_\varepsilon V=\delta z$, and in $\eta$ gives $\partial_\eta V=\dot x$. Since $V_{\eta_c}'(0)=0$, we obtain $x_c'(0)=u_c'(0)\,e_c+\eta_c'(0)\,\dot x+\delta z$, and $\nabla^3\varphi=\kappa_3/\eta^3$ (~\cref{lem:cumulants}) gives \eqref{eq:foldshift-x}.

\subsection{Proof of~\cref{prop:cusp}}
\label{app:prop:cusp}
As in the proof of Proposition~\ref{prop:foldpersist}, the perturbed reduced function satisfies
\[
  \Phi_{\eta,\varepsilon}(u)=\Phi_\eta(u)+\varepsilon\bigl(h+O(|u|+|\chi|)\bigr)+O(\varepsilon^2),
\]
and, by \eqref{eq:Phi-joint} with $A=B=0$ and by the oddness of $\Phi_\eta$ (~\cref{lem:sym}), $\Phi_\eta(u)=D\chi u+\frac C6u^3+O(|\chi|u^3+\chi^2|u|+|u|^5)$. With the scaling $u=|\varepsilon|^{1/3}U$, $\chi=|\varepsilon|^{2/3}X$, and $\sigma:=\operatorname{sgn}\varepsilon$, we get
\[
  |\varepsilon|^{-1}\Phi_{\eta,\varepsilon}(u)=\sigma h+DXU+\frac C6U^3+O(|\varepsilon|^{1/3}),
\]
with a remainder that is smooth in $(U,X,|\varepsilon|^{1/3})$. The fold conditions $\Phi=\partial_U\Phi=0$ for the leading part read $\sigma h+DXU+\frac C6U^3=0$ and $DX+\frac C2U^2=0$. Eliminating $X$ gives $\sigma h-\frac C3U^3=0$, hence $U^3=3\sigma h/C$, $X=-CU^2/(2D)$, and $X^3=-9Ch^2/(8D^3)$, $U=-3\sigma h/(2DX)$. The Jacobian of the leading system with respect to $(U,X)$ at this solution is $\bigl(\begin{smallmatrix}0&DU\\ CU&D\end{smallmatrix}\bigr)$, with determinant $-CDU^2\neq0$ because $U\neq0$. The implicit function theorem in the variable $|\varepsilon|^{1/3}$ gives the fold and \eqref{eq:cusp}. If $C<0$ and $D<0$, then $-9C/(8D^3)<0$, so $\chi<0$. Finally, for $\chi>0$ the perturbed reduced equation has a single zero near $u=0$, $u\simeq-\varepsilon h/(D\chi)$, which has the sign of $\varepsilon h$ when $D<0$; this zero is continued through the speciation, while the second mode is born at the fold on the opposite side.

\section{Neural networks for score training}
\crefalias{section}{appendix}
\label{sec:NN}
We consider 6 different small fully connected neural networks.
Each network:
\begin{itemize}
\item takes as input the rescaled position $x/\sqrt{\eta+\sigma_{\rm data}^2}$, where $\sigma_{\rm data}^2$ is the average per-coordinate variance of the pool, together with the features $t$, $\sin(\pi kt)$, $\cos(\pi kt)$, $k=1,2,3$, of $t:=\log(\eta/\eta_c)/\log3$
\item is composed of three hidden layers of equal width with SiLU activations. The considered widths are 64, 128, 256
\item has output $\hat\varepsilon_\theta(x,\eta)$ which defines the score $S_\theta(x,\eta):=-\hat\varepsilon_\theta(x,\eta)/\sqrt\eta$. 
\item is trained on denoising score matching, i.e.\ by minimizing $\mathbb{E}\,\|\hat\varepsilon_\theta(y+\sqrt\eta\,z,\eta)-z\|^2$ with $y$ drawn from the pool, $z\sim\mathcal N(0,I_d)$ and $\log\eta$ uniform on $[\log(\eta_c/3),\log(3\eta_c)]$, with Adam, batches of $512$ samples and a cosine-decaying learning rate starting at $2\cdot10^{-3}$. Training is restricted to noise levels around $\eta_c$ because we are interested in the local behavior. Different numbers of weight updating are considered as specified in \cref{tab:fmnist-learned}.

\end{itemize}

\end{document}